\documentclass[10pt,twocolumn,letterpaper]{article}

\usepackage[pagenumbers]{cvpr}

\usepackage{iftex}
\ifXeTeX\usepackage[T1]{fontenc}\fi

\usepackage{amsthm}%
\newtheorem{proposition}{Proposition}
\usepackage{pifont}%
\definecolor{tabyes}{rgb}{0.10,0.50,0.22}%
\definecolor{tabno}{rgb}{0.80,0.12,0.12}%
\definecolor{tablocal}{gray}{0.45}%
\usepackage{tikz}%
\usetikzlibrary{positioning}

\usepackage{currfile}%

\usepackage{caption}%

\newcommand{\TODO}[1]{\textbf{\color{red}[TODO: #1]}}
\newcommand{\ph}[1]{{\color{red}#1}}%
\usepackage{multirow}
\newcommand{\ci}[2]{{\scriptsize\color{black!55}\,[#1,\,#2]}}
\usepackage{colortbl}
\definecolor{oursrow}{gray}{0.90}

\newsavebox{\resfitbox}
\newcommand{\resfit}[2]{\sbox{\resfitbox}{#2}\ifdim\wd\resfitbox>#1\relax\resizebox{#1}{!}{\usebox{\resfitbox}}\else\usebox{\resfitbox}\fi}

\newcommand{\resContactErr}{14.6}

\newcommand{\resWithheldTA}{23.0}
\newcommand{\resWithheldVI}{27.4}

\newcommand{\resWithheldHawor}{37.8}

\newcommand{\resRealOccTA}{86}
\newcommand{\resRealOccVI}{149}

\newcommand{\resRealOccHawor}{238}

\newcommand{\resSeenWithheldTA}{3.6}
\newcommand{\resSeenWithheldVI}{16.2}

\newcommand{\resUnseenWithheldTA}{6.7}
\newcommand{\resUnseenWithheldVI}{11.4}

\newcommand{\resUnreliableShare}{24}

\newcommand{\resSweepTwoTA}{20.6}
\newcommand{\resSweepTwoVI}{28.6}

\newcommand{\resSweepFiveTA}{34.3}
\newcommand{\resSweepFiveVI}{86.7}
\newcommand{\resSweepFiveReduction}{60}

\newcommand{\resSweepTenTA}{62.6}
\newcommand{\resSweepTenVI}{164.7}

\newcommand{\resWithheldContactTA}{18.9}
\newcommand{\resWithheldContactVI}{33.6}

\newcommand{\resProjSeen}{5.2}
\newcommand{\resProjUnseen}{7.5}
\newcommand{\resProjRealOcc}{51.1}
\newcommand{\resWtwoTargetAll}{14.8}

\newcommand{\resWtwoTargetWithheldTA}{25.1}
\newcommand{\resWtwoTargetWithheldVI}{35.6}

\newcommand{\resRealOccScanTA}{14.4}
\newcommand{\resRealOccScanVIfilter}{24.9}
\newcommand{\resRealOccScanVI}{51.1}

\newcommand{\resAlongShortTA}{6}
\newcommand{\resAlongShortVI}{4}
\newcommand{\resAlongLongTA}{15}
\newcommand{\resAlongLongVI}{95}
\newcommand{\resSweepShortS}{0.6}
\newcommand{\resAblFiveVIThreeD}{86.7}
\newcommand{\resAblFiveSurfThreeD}{84.8}
\newcommand{\resAblFiveNoSlipThreeD}{38.8}
\newcommand{\resAblFiveTAThreeD}{34.3}
\newcommand{\resAblFiveVIAlong}{33.3}
\newcommand{\resAblFiveSurfAlong}{14.7}

\newcommand{\resAblFiveTAAlong}{10.4}
\newcommand{\resWeakTA}{24.4}
\newcommand{\resWeakOK}{30.4}

\newcommand{\resWindowL}{2}
\newcommand{\resWithheldWindowMax}{10}

\renewcommand{\TODO}[1]{}
\renewcommand{\ph}[1]{#1}

\definecolor{cvprblue}{rgb}{0.21,0.49,0.74}
\usepackage[pagebackref,breaklinks,colorlinks,allcolors=cvprblue]{hyperref}

\title{TouchAtlas: World-Registered Tactile Maps with a Wearable Glove}

\author{Gongbo Yang$^{1}$ \enspace Saeid Soltanian$^{1}$ \enspace Jay Chung$^{2}$ \enspace Amir Servati$^{1}$ \enspace Hyejeong Choi$^{1}$ \enspace Peyman Servati$^{1,\dagger}$\\
$^{1}$University of British Columbia \qquad $^{2}$Texavie\\
{\tt\small gongboyang@ece.ubc.ca \qquad $^{\dagger}$peymans@ece.ubc.ca}}

\begin{document}
\twocolumn[{%
\renewcommand\twocolumn[1][]{#1}%
\maketitle
\begin{center}
\includegraphics[width=\linewidth]{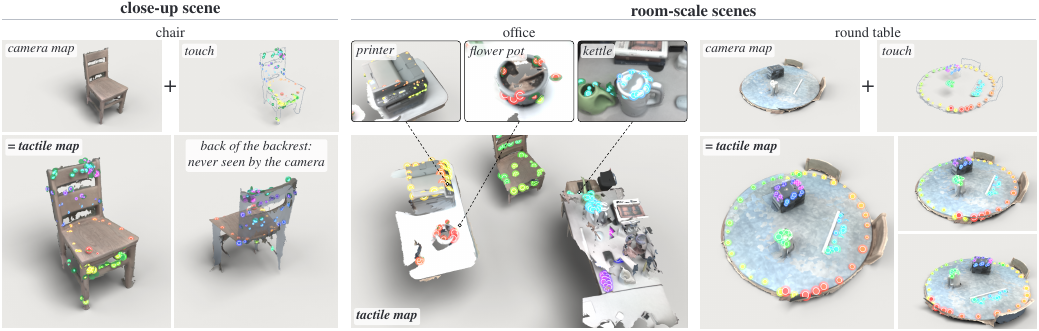}
\captionsetup{hypcap=false}
\captionof{figure}{\textbf{World-registered tactile maps.}
Left: a chair. The visual map from the head camera and the contacts registered by TouchAtlas together form its tactile map.
The contacts on the back of the backrest lie on a face the camera never saw.
Middle: an office setup, with close-ups of three touched objects.
Right: a round table.
Contacts are coloured by the order of touch and sized by pressure. Green discs are the touch patches fitted to them.}
\captionsetup{hypcap=true}
\label{fig:\currfilebase}%
\end{center}

}]
\begin{abstract}
Tactile gloves capture how pressure is distributed over the hand, but not where in the scene the contact took place.
We present TouchAtlas, which builds \emph{world-registered tactile maps} from tactile gloves and egocentric RGB-D cameras.
Each contact is registered where and when it occurred, with its pressure distribution and the hand region that made it.
Registering touch is hardest exactly when it happens.
As the hand presses and reaches, the contact is hidden from the camera,
and the hand pose estimated from vision becomes unreliable or unavailable.
The touched surfaces are often hidden too, behind or under the objects the hand reaches around.
Our key observation is that the dependence runs both ways:
the hand pose places a contact in the world, and the contact in turn constrains the hand pose.
First, a contact must lie on the surface it touches.
The camera's map gives this surface where the camera has seen it, and repeated contacts reveal it where the camera has not.
Second, a contact that is held stays at one point.
TouchAtlas combines both constraints with the glove's inertial orientation and intermittent visual observations
to jointly estimate the hand trajectory and the touched surfaces in the world frame.
The resulting map completes what the camera sees with the surfaces only the hand has reached,
and a contact on a mapped surface is registered on that surface even when the hand is not seen.
It runs online at 25\,Hz and requires no external tracking.
When the visual hand pose is unavailable, TouchAtlas reduces the median distance of contacts to a separate scan of the scene
from \ph{\resSeenWithheldVI} to \ph{\resSeenWithheldTA}\,mm ($-$\ph{78}\%) on surfaces the camera has seen,
and from \ph{\resUnseenWithheldVI} to \ph{\resUnseenWithheldTA}\,mm ($-$\ph{41}\%) on surfaces it has not, relative to visual--inertial tracking.
The error of a contact on a known target drops from \ph{\resWithheldContactVI} to \ph{\resWithheldContactTA}\,mm ($-$\ph{44}\%).
\end{abstract}

\section{Introduction}
\label{sec:intro}

Our hands act on the world through contact,
yet in today's egocentric systems the head-mounted camera mostly sees the back of the hand, while contact happens on the palm and fingertip side.
Touch senses what vision misses: whether the hand is actually in contact, which exact part of it is touching, and how pressure is distributed across it.
Tactile gloves now measure this signal over the whole hand~\cite{sundaram2019stag,jiang2024vitam,song2025opentouch},
and recent datasets record it alongside video at scale~\cite{li2026touchscale,zhang2026glove2hand,zhou2026touchanything}.
However, in these systems touch remains a signal on the hand, recorded in the sensor's own frame.
It is used to identify objects, predict contact, or supervise vision, but it is not registered in the scene.
We cannot ask where on the table the hand pressed, which side of the box the fingers held,
or which surfaces in a room are touched most often and by which part of the hand.
Conversely, the camera maps only the surfaces it sees, not those the hand reaches behind or under objects.

\begin{figure*}[t]
\centering
\includegraphics[width=\textwidth]{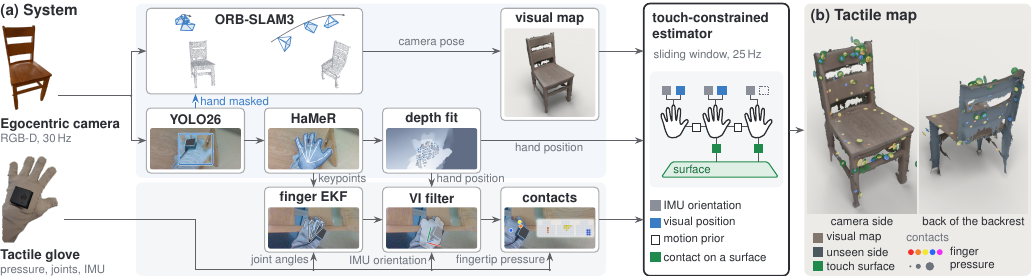}
\caption{\textbf{TouchAtlas.} (a) System. ORB-SLAM3 tracks the egocentric RGB-D camera with the hand masked and defines the world frame.
The depth images fused along its path give the visual map.
YOLO26 detects the glove, HaMeR reconstructs the hand, and the hand position is fitted to the depth of the glove fabric.
The VI filter fuses this position with the IMU orientation and the finger EKF. Contacts come from the fingertip pressure.
The touch-constrained estimator refines the hand trajectory over a sliding window.
It places each contact on the visual map, or on the touch surface where the camera saw none,
even without a visual hand position (dashed square).
(b) The resulting tactile map of a chair, from the camera's side and from behind.
It shows the visual map and its unseen side, the touch surface, and the registered contacts (colour: finger; size: pressure).}
\label{fig:system}
\end{figure*}
We introduce \textbf{TouchAtlas}, a method that builds \emph{world-registered tactile maps} from wearable sensors alone:
tactile gloves and egocentric RGB-D cameras.
In a tactile map, each contact is registered at the point it touched on a surface
and carries its pressure distribution, the exact hand region that made it, and its timestamp.
The trajectory of the hand links these contacts over time.
Where the camera has mapped the scene, the contacts lie on its surfaces.
Where it has not, the contacts themselves reveal the touched surfaces,
so the map completes the camera's view with the surfaces only the hand has reached.
It turns transient touch into a persistent account of how a scene was physically handled.
\cref{fig:teaser} shows such maps of a chair, an office setup and a round table.
On the chair, the contacts on the back of the backrest lie on a face the camera never saw.
To our knowledge, TouchAtlas is the first framework to register the contacts of a tactile glove in a world frame with wearable sensors,
and the first to use these contacts to constrain the position of the hand.

Registering touch in the world is harder than it appears.
A taxel, a single sensing element of the glove, reaches the world frame through a chain of estimates:
the pose of the camera, the pose of the hand relative to the camera, the finger kinematics,
and the offset from the finger to its sensing pad.
The weakest link is the pose of the hand relative to the camera, and it usually fails at the very moment of contact.
While touching, the hand is curled, partly hidden behind the touched object, or near the edge of the view,
so visual hand pose estimates degrade or vanish.
In our recordings, \ph{\resUnreliableShare\%} of contact events occur while the visual hand pose is unavailable or unreliable.
The glove's IMU keeps updating the hand orientation through these gaps, but not its position, which drifts quickly when integrated.
The touched surfaces are often hidden too, behind or under the objects the hand reaches around,
where the camera's map has no surface to place the contacts on.

Our key observation is that this dependence runs both ways:
the pose of the hand places a contact in the world, and the contact in turn constrains where the hand is.
First, a contact always lies on a surface.
If the camera has seen that surface, the visual map already contains it.
If not, repeated contacts on a static object all fall on the same surface, so the contacts themselves reveal it.
Either way, the surface pins the distance of the hand to it.
Second, a contact that is held stays at one point,
which ties the position of the hand to that point for as long as the contact lasts.
Because the touched surfaces are kept in the map, a later contact on one of them is placed on it,
much like a loop closure in SLAM.
TouchAtlas combines these constraints with the glove's inertial orientation and intermittent visual observations
to jointly estimate the hand trajectory and the touched surfaces.
Contacts thus stay registered when vision fails, and the surfaces the camera never saw enter the map.

Prior work localises the hand without touch or uses touch to estimate pose only locally (\cref{tab:prior}).
Contacts in the world frame have been used for feet rather than hands.
TouchAtlas brings touch into the estimation of a human hand's pose in a scene-level world frame,
and maps the touched surfaces, including those no camera saw.

We realise TouchAtlas with a tactile glove that senses pressure at the fingertips, finger joint angles and hand orientation,
worn together with an egocentric RGB-D camera. It runs online at 25\,Hz.
When the visual hand pose is unavailable, TouchAtlas reduces the median distance of contacts to a separate scan of the scene
from \ph{\resSeenWithheldVI} to \ph{\resSeenWithheldTA}\,mm on surfaces the camera has seen,
and from \ph{\resUnseenWithheldVI} to \ph{\resUnseenWithheldTA}\,mm on surfaces it has not, relative to visual--inertial tracking.
The error of a contact on a known target drops from \ph{\resWithheldContactVI} to \ph{\resWithheldContactTA}\,mm.

In summary, our contributions are:
\begin{itemize}
  \item \textbf{World-registered tactile maps.}
        Each contact is registered in the world frame on the surface it touched,
        with its pressure distribution and hand region over time, together with the hand trajectory.
  \item \textbf{Touch completes the visual map.}
        Contacts are registered on the surfaces the camera has seen, and themselves locate the surfaces it has not.
  \item \textbf{Touch as a constraint on hand pose.}
        An online estimator fuses contacts with inertial orientation and intermittent vision,
        and we show which directions of the hand position each contact constraint fixes.
\end{itemize}

\section{Related Work}
\label{sec:related}

\paragraph{Touch on the hand.}
Tactile gloves measure pressure over the hand with piezoresistive or stretchable sensor arrays~\cite{sundaram2019stag,jiang2024vitam,ruppel2024elasticglove,lin2026artglove},
often together with finger joint angles or hand pose~\cite{liu2017poseforceglove,ruppel2024elasticglove,lin2026artglove}.
Recent datasets pair full-hand touch with egocentric video at scale~\cite{li2026touchscale,song2025opentouch,zhou2026touchanything},
and use it to predict contact, to synthesise bare-hand video~\cite{zhang2026glove2hand},
or to supervise models that infer touch from vision alone~\cite{zhang2026egotac,zhou2026touchsight,yang2026ego2contact}.
Touch has been located on surfaces without a glove, but only on prepared ones.
Thermal imprints give a contact map in the frame of one object, without pressure or time~\cite{brahmbhatt2019contactdb,brahmbhatt2020contactpose},
and instrumented pads measure pressure on a single plane~\cite{grady2022pressurevision,zhao2025egopressure}.
Otherwise, touch is expressed in the frame of the hand or the sensor and is not registered in the scene.

\paragraph{Localising the gloved hand.}
Registering touch requires the pose of the hand in a common frame.
Tactile gloves have been localised with optical motion capture or base-station trackers~\cite{ni2026tactidex,qin2026dextouchwm,zhou2026touchanything},
which is accurate but ties capture to an instrumented space.
TactiDex~\cite{ni2026tactidex} also pulls the fingers in contact onto the object surface, but keeps the wrist pose as tracked.
Wearable alternatives fuse a head camera with glove IMUs~\cite{choi2026robusthand,kou2026avihit} or read fiducials on the glove and around the workspace~\cite{shin2026headcamera}.
Without a glove, HaWoR~\cite{zhang2025hawor} and Dyn-HaMR~\cite{yu2025dynhamr} reconstruct hand motion in a world frame from a moving camera,
with motion priors bridging the frames in which the hand is not observed.
In all of these, the hand pose comes from tracking alone. Touch, where recorded, is placed with that pose but does not improve it,
so nothing measures the position of a hand that is hidden while it touches.
Our visual--inertial baseline represents this class, and we compare with HaWoR directly.

\begin{figure*}[t]
\centering
\includegraphics[width=\textwidth]{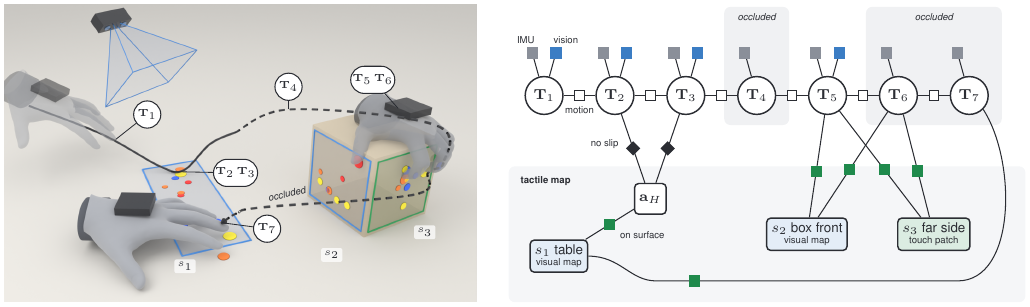}
\caption{Factor graph of the estimator on a schematic sequence (left).
The hand is seen and holds a contact on the table ($\mathbf{T}_2$, $\mathbf{T}_3$). No-slip factors tie both poses to the anchor $\mathbf{a}_H$, which lies on the table surface $s_1$.
The table and the front of the box ($s_1$, $s_2$) come from the camera's map. The far side of the box ($s_3$) is never seen by the camera.
Reaching around the box with the back of the hand in view ($\mathbf{T}_5$), the fingers touch $s_3$ and create a touch patch there.
When the hand is occluded ($\mathbf{T}_4$, $\mathbf{T}_6$, $\mathbf{T}_7$), only the IMU orientation and the motion prior remain.
The contacts of the grasp ($\mathbf{T}_6$) then lie on $s_2$ and on the patch $s_3$.
Touching the table again ($\mathbf{T}_7$) places the contact on $s_1$ without a visual hand pose.
Contacts are coloured by finger and sized by pressure, as in \cref{fig:system}.}
\label{fig:factor_graph}
\end{figure*}

\paragraph{Touch in pose estimation.}
Touch has also been used to estimate pose, but relative to an object or a sensor.
ViTaM-D~\cite{yu2025vitamd} refines the hand pose against a held object with a force-aware contact energy.
Tactile SLAM estimates the pose and shape of an object from contact, with tactile sensors on robots
whose kinematics locate them~\cite{suresh2021tactileslam,huang2026gelslam,suresh2024neuralfeels}.
Our estimator shares their principle of estimating pose and an unknown surface together online,
but for a human hand wearing a low-resolution pressure glove, and at the scale of a scene rather than a single object.
Touch has been registered in a scene-level frame with optical tactile sensors
whose pose a mounted camera or a robot provides~\cite{dou2024tarf,swann2024touchgs}. There, touch refines the scene, not the pose.
Closest in output, TouchMap-OR~\cite{ktistakis2026touchmapor} reconstructs where clinicians touched surfaces in operating rooms
from fixed multi-view RGB-D, inferring contacts from hand--surface proximity rather than sensing them.

\newcommand{\yes}{\textcolor{tabyes}{\ding{51}}}
\newcommand{\no}{\textcolor{tabno}{\ding{55}}}
\newcommand{\local}{\textcolor{tablocal}{local}}
\newcommand{\obj}{\textcolor{tablocal}{object}}
\begin{table}[b]
\centering
\small
\setlength{\tabcolsep}{2.5pt}
\resizebox{\columnwidth}{!}{%
\begin{tabular}{@{}lccccc@{}}
\toprule
 & Human & World & Touch & & Unseen \\
 & hand & frame & in pose & Map & surfaces \\
\midrule
Gloves + ext.\ tracking~\cite{ni2026tactidex,qin2026dextouchwm,zhou2026touchanything} & \yes & \yes & \local & \no & \no \\
Visual--inertial tracking~\cite{choi2026robusthand,kou2026avihit} & \yes & \yes & \no & \no & \no \\
World-space hand recon.~\cite{zhang2025hawor,yu2025dynhamr} & \yes & \yes & \no & \no & \no \\
Hand--object recon.~\cite{jiang2024vitam,yu2025vitamd} & \yes & \no & \local & \no & \local \\
Tactile SLAM~\cite{huang2026gelslam,suresh2021tactileslam,suresh2024neuralfeels} & \no & \obj & \local & \yes & \yes \\
Touch in a 3D scene~\cite{swann2024touchgs} & \no & \yes & \no & \yes & \yes \\
Touch from proximity~\cite{ktistakis2026touchmapor} & \yes & \yes & \no & \yes & \no \\
\midrule
\textbf{TouchAtlas (ours)} & \yes & \yes & \yes & \yes & \yes \\
\bottomrule
\end{tabular}}
\caption{Comparison with prior work by class.
\emph{Human hand}: not a robot or a hand-held sensor.
\emph{World frame}: scene-level (object: the frame of one object).
\emph{Touch in pose}: touch constrains the pose in the world frame (local: only relative to a held object or the sensor).
\emph{Map}: a persistent map of touched surfaces.
\emph{Unseen surfaces}: touch is registered on surfaces the camera does not see.}
\label{tab:prior}
\end{table}
\paragraph{Contacts in localisation.}
Contacts also constrain where a body is in the world.
Foot-mounted inertial navigation resets the velocity to zero while the foot is on the ground~\cite{foxlin2005pedestrian},
and legged robots hold the stance foot fixed while fusing leg kinematics with an IMU~\cite{bloesch2012legged,hartley2020contact}.
Human motion capture from body-worn sensors uses foot contacts in the same way~\cite{guzov2021hps,yi2023egolocate,zhang2024mmvp},
and IMU-HOI~\cite{lin2026imuhoi} infers hand--object contacts from IMUs to refine the body pose.
Our no-slip constraint is the hand counterpart of these stance constraints, which hold a contact in place while it lasts.
TouchAtlas additionally maps the touched surfaces without a prior scan, from the camera where it sees them and from contact where it does not.
A later contact on such a surface is then registered on it without vision.

\section{System}
\label{sec:system}

\paragraph{Sensors.}
TouchAtlas runs on a tactile glove and an RGB-D camera worn on the head (\cref{fig:system}).
The glove carries a pressure array of 48 taxels on each fingertip, joint-angle sensing on each finger,
and a 9-axis IMU on the back of the hand.
Each array has 8 columns and 7 rows, with four taxels left out at each upper corner so that it wraps around the tip.
Fingertip arrays stream at 80\,Hz and the IMU orientation at 200\,Hz over a wired link.
The estimator uses the fingertips.
The camera, a RealSense D435, records $848\times480$ colour and aligned depth at 30\,Hz.
On an RTX 4070 Ti SUPER, camera tracking runs at 30\,Hz and hand estimation at 25\,Hz
with the hand in view (\cref{sec:app_impl}).

\paragraph{Hand tracking.}
ORB-SLAM3~\cite{campos2021orbslam3} in RGB-D mode tracks every camera frame, with the region around the hand masked out,
and its map defines the world frame $\mathcal{W}$ in which contacts are registered.
Off-the-shelf hand detectors rarely fire on the sensor glove, so we fine-tune YOLO26n~\cite{jocher2026yolo26} on images of the glove.
HaMeR~\cite{pavlakos2024reconstructing} reconstructs a MANO hand from the detected crop.
HaMeR's depth estimate is unreliable for the glove, so we take the hand position from the depth image.
The enclosure box on the back of the hand, its cable and folds of the fabric stand more than 12\,mm proud of the hand
and cover about \ph{60\%} of the depth samples on the palm.
Aligning the HaMeR palm to all of them would place the hand too close to the camera.
We therefore fit the palm to the glove fabric, and accept the fit only when it agrees with the HaMeR palm (\cref{sec:app_impl}).
The hand orientation comes from the glove IMU, aligned online to the world frame by fitting the IMU-to-world and hand-to-IMU rotations
to visually valid frames.
The hand position is refitted with the IMU orientation when the HaMeR orientation is inconsistent,
and is filtered by a Kalman filter with a decaying-velocity model.
Each finger is estimated by an EKF that fuses the glove joint angles with HaMeR keypoints.
It learns the bias of the glove angles while the finger is visible and follows the glove alone when it is occluded or in contact.
This fusion is a causal visual--inertial filter (VI filter). Without a visual observation, it holds the last hand position.
The touch-constrained estimator of \cref{sec:method} refines its trajectory and registers the contacts.
\Cref{sec:experiments} compares it against the VI filter and against the estimator with the touch constraints removed.

\section{World-Registered Tactile Mapping}
\label{sec:method}

\begin{figure*}[t]
\centering
\includegraphics[width=\textwidth]{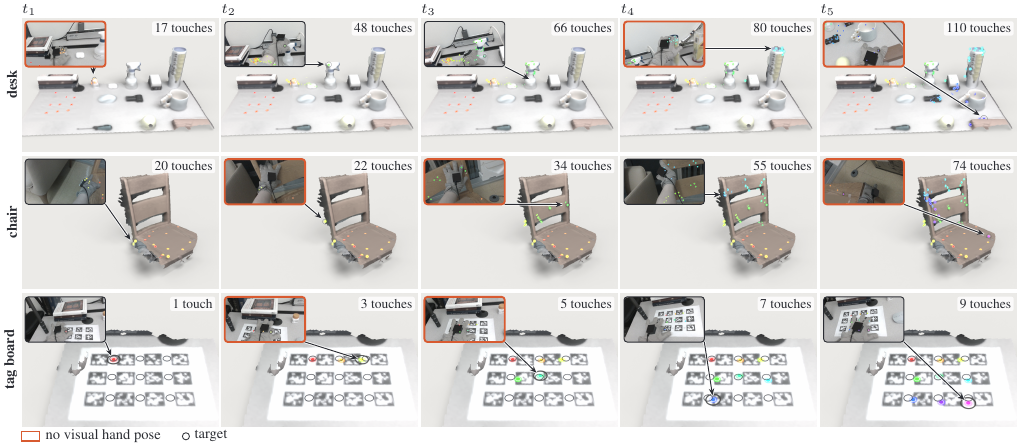}
\caption{\textbf{The map-building process.} Each tile is the tactile map after the stated number of touches, with contacts coloured by the order of touch.
The inset is the head-camera frame at the latest press, and the arrow marks that press on the map.
A red inset means that there was no visual hand pose at that press.
Top: a desk with everyday objects. Middle: a chair. Bottom: nine presses on the known targets of a tag board (rings), with the contacts drawn on the board plane.}
\label{fig:results_grid}
\end{figure*}

We first define the tactile map and the contacts it is built from (\cref{sec:map}),
then the two touch constraints (\cref{sec:constraints}),
and finally the online estimator that combines them with inertial and visual observations (\cref{sec:estimator}, \cref{fig:factor_graph}).

\subsection{Contacts and the tactile map}
\label{sec:map}

Let $\mathbf{R}_t \in SO(3)$ and $\mathbf{p}_t \in \mathbb{R}^3$ be the orientation and position of the hand, taken at the palm, in the world frame $\mathcal{W}$ at time $t$,
and $\boldsymbol{\theta}_t$ the finger joint angles from the finger EKF (\cref{sec:system}).
The orientation $\mathbf{R}_t$ is taken from the IMU, aligned to the world frame as in \cref{sec:system}. The estimator solves for the position $\mathbf{p}_t$.
A fingertip pad $r$ is in contact when the sum of its pressure readings exceeds a threshold $\tau_\text{on}$.
Its contact point in the hand frame is
\begin{equation}
  \mathbf{q}_r(t) = g_r(\boldsymbol{\theta}_t) + \mathbf{c}_r ,
  \label{eq:pad}
\end{equation}
where $g_r$ is a fixed point on the pad of the distal phalanx, obtained by forward kinematics,
and $\mathbf{c}_r$ is a constant correction that absorbs the bias of the glove's finger kinematics during a press.
It is calibrated per finger at the start of a session from presses on known targets (\cref{sec:setup}).
A contact at time $t$ is placed in the world at
\begin{equation}
  \mathbf{x}_r(t) = \mathbf{R}_t\,\mathbf{q}_r(t) + \mathbf{p}_t .
  \label{eq:register}
\end{equation}
The pressure distribution of the pad is stored with the contact.
A \emph{contact event} is a run of at least three consecutive frames in which a pad is in contact.

The tactile map $\mathcal{M}$ holds the contacts $c = (\mathbf{x}_c, \mathbf{P}_c, r_c, t_c)$,
with world position, pressure distribution over the taxels of the pad, hand region and time;
the hand trajectory $\{(\mathbf{R}_t, \mathbf{p}_t)\}$;
and the surfaces that support the contacts.
A surface $\mathcal{S}$ answers one query: for a point $\mathbf{x}$ it returns the closest surface point $\boldsymbol{\pi}_\mathcal{S}(\mathbf{x})$
and the unit normal $\mathbf{n}_\mathcal{S}(\mathbf{x})$ there.
The map has two surfaces.
The \emph{visual map} $\mathcal{S}_V$ is a mesh fused from the depth images of the head camera with the hand masked out.
Contacts do not change it.
The \emph{touch surface} $\mathcal{S}_T$ covers what the camera has not seen.
It consists of local tangent patches $s = (\mathbf{o}_s, \mathbf{n}_s)$, a centre and a normal, fitted to the contacts that do not lie on the visual map.
Only contacts made while the hand is observed by the camera create patches and move them.
This is the common case on a surface the camera cannot see.
The fingers reach around an edge and touch its far side, while the back of the hand stays in view.
No model of the scene or of the objects is given. \Cref{sec:app_impl} details both surfaces.

\subsection{Touch constraints}
\label{sec:constraints}

\paragraph{Contact on a surface.}
A contact must lie on the surface it touches.
For a contact of pad $r$ at time $t$ associated with a surface $\mathcal{S} \in \{\mathcal{S}_V, \mathcal{S}_T\}$,
\begin{equation}
  e_\text{surf} = \mathbf{n}_\mathcal{S}(\mathbf{x})^\top \big(\mathbf{x} - \boldsymbol{\pi}_\mathcal{S}(\mathbf{x})\big),
  \qquad \mathbf{x} = \mathbf{R}_t\,\mathbf{q}_r(t) + \mathbf{p}_t ,
  \label{eq:surf}
\end{equation}
the signed distance of the contact to the surface.
\Cref{eq:surf} constrains one degree of freedom of the hand position, its distance to the surface.
A contact event is associated with the visual map if at least half of its frames lie within $\epsilon_d$ of it
and the surface normal is within 70$^\circ$ of the nearest opposed normal of the fingertip skin, from the pad to the tip.
We set $\epsilon_d = 3$\,cm when the hand is observed by the camera and 10\,cm when it is not.
An event that fails these tests is tested against the touch surface in the same way, and otherwise has no surface constraint.
The test covers the whole fingertip because small objects are touched with the tip as often as with the pad.
In our recordings, the surface normal lies \ph{38--83}$^\circ$ from the pad normal along this arc.
For the same reason, we do not constrain the pad normal to oppose the surface normal.
\paragraph{No slip.}
While a contact is held, the touched point does not move in the world.
Each contact event $H$ receives an anchor $\mathbf{a}_H \in \mathbb{R}^3$, and for every frame $t \in H$
\begin{equation}
  \mathbf{e}_\text{slip} = \mathbf{R}_t\,\mathbf{q}_r(t) + \mathbf{p}_t - \mathbf{a}_H .
  \label{eq:slip}
\end{equation}
Given the orientation $\mathbf{R}_t$ and the finger kinematics, \cref{eq:slip} fixes the hand position to
$\mathbf{p}_t = \mathbf{a}_H - \mathbf{R}_t\,\mathbf{q}_r(t)$, all three degrees of freedom, for as long as the contact lasts.
It does not locate the anchor itself.
The anchor is placed by visual observations before or after the contact, by the surface it lies on, or by both.
Between contacts and without vision, only the motion prior constrains the position.
We do not detect slip explicitly. A robust loss limits the influence of a finger that slides or rolls.

\paragraph{What touch constrains.}
With the orientation and the finger kinematics given, both constraints act on the hand position $\mathbf{p}_t$.
What they determine without vision can be stated exactly.

\begin{proposition}
\label{prop:touch}
Let $\mathbf{R}_t$ and $\mathbf{q}_r(t)$ be given, and consider frames in which the hand is not observed by the camera.
\begin{enumerate}
  \item[(i)] \emph{Surface.}
        A contact on a mapped surface with unit normal $\mathbf{n}$ determines $\mathbf{n}^\top \mathbf{p}_t$.
        Contacts on surfaces whose normals span $k$ dimensions determine $k$ components of the position and leave $3-k$ free.
  \item[(ii)] \emph{No slip.}
        Within a contact event $H$, \cref{eq:slip} determines $\mathbf{p}_t - \mathbf{p}_{t'}$ for all $t, t' \in H$,
        and is unchanged by a common translation of the positions $\{\mathbf{p}_t\}_{t \in H}$ and the anchor $\mathbf{a}_H$.
  \item[(iii)] \emph{Touch surface.}
        If a touch patch is supported only by contacts in these frames,
        \cref{eq:surf,eq:slip} are unchanged by a common translation of the patch, the anchors and the positions of these frames.
\end{enumerate}
\end{proposition}

Each statement follows from the form of the residuals (\cref{sec:app_impl}).
The statements explain three choices of the method.
By (i), on a single plane the motion along the plane remains free between contacts.
Touching surfaces with three independent normals, such as a table and two sides of a box ($s_1$--$s_3$ in \cref{fig:factor_graph}), determines the full position.
By (ii), a held contact keeps the hand in place but does not say where, so every anchor needs a visual observation or a surface.
By (iii), a surface the camera has not seen constrains the hand only after vision has located a contact on it.
This is why only contacts made while the hand is observed create touch patches (\cref{sec:map}).
Once a surface is located, by the camera or by such contacts, a contact made on it without vision
is registered on that surface ($\mathbf{T}_7$ in \cref{fig:factor_graph}).
\Cref{sec:ablations} tests each statement.

\subsection{Online estimation}
\label{sec:estimator}

The estimator keeps a window of the most recent hand positions $\{\mathbf{p}_t\}_{t \in \mathcal{T}}$, one per frame over the last $L$ seconds,
together with the anchors of the contact events in the window,
and minimises
\begin{equation}
\begin{aligned}
  E = {} & \sum_{t \in \mathcal{V}} \rho\big( \| \mathbf{p}_t - \hat{\mathbf{p}}_t \|^2 / \sigma_V^2 \big)
          + \sum_{\text{contacts}} \rho\big( e_\text{surf}^2 / \sigma_S^2 \big) \\
        & + \sum_{t} \rho\big( \| \mathbf{p}_{t+1} - 2\mathbf{p}_t + \mathbf{p}_{t-1} \|^2 / \sigma_{A,t}^2 \big) \\
        & + \sum_{H} \sum_{t \in H} \rho\big( \| \mathbf{e}_\text{slip} \|^2 / \sigma_H^2 \big)
          + E_\text{prior} .
\end{aligned}
\label{eq:energy}
\end{equation}
The first term ties the position to the visual hand position $\hat{\mathbf{p}}_t$ on the frames $\mathcal{V}$ where the depth fit of \cref{sec:system} is valid.
The second and fourth are the touch constraints.
The third is a constant-velocity prior, the only position term when neither vision nor touch is available.
$\rho$ is a Huber loss that limits the influence of wrong detections, wrong surface associations and undetected slip.
A position that leaves the window is marginalised into $E_\text{prior}$.

We use $L = \ph{\resWindowL}$\,s.
\Cref{sec:app_impl} describes how the surfaces are updated with the trajectory and when a contact is final,
and gives the noise parameters, the solver and its timing.

\section{Experiments}
\label{sec:experiments}

\begin{table}[t]
\centering
\footnotesize\setlength{\tabcolsep}{3pt}\resfit{\columnwidth}{\begin{tabular}{@{}lccc@{}}
\toprule
Median & Contact & \multicolumn{2}{c}{Contact to scan, no visual pose} \\
\cmidrule(lr){3-4}
{\scriptsize\color{black!55}[95\% interval]}, mm & to target & seen surface & unseen surface \\
\midrule
VI filter & 16.7\ci{12.5}{20.9} & 18.1\ci{15.3}{19.6} & 15.1\ci{9.0}{20.8} \\
VI smoother & 16.0\ci{11.5}{21.6} & 16.2\ci{13.7}{20.5} & 11.4\ci{8.6}{17.4} \\
VI + map projection & 15.5\ci{13.8}{17.5} & 5.2\ci{3.2}{7.8} & 7.5\ci{5.6}{14.0} \\
\rowcolor{oursrow}TouchAtlas (ours) & 14.6\ci{13.0}{17.2} & \textbf{3.6}\ci{3.0}{4.7} & \textbf{6.7}\ci{4.3}{9.2} \\
\midrule
$\Delta$ (paired) & $-$1.4\ci{$-$6.2}{$+$2.0} & $-$12.5\ci{$-$16.8}{$-$10.0} & $-$4.7\ci{$-$11.3}{$-$1.9} \\
\bottomrule
\end{tabular}
}
\caption{\textbf{Map accuracy.}
Left: distance between a registered contact and the known target it was made on, with all vision available.
Right: distance between a contact registered without a visual hand pose and the scan of the scene,
on surfaces the head camera saw and did not see.
$\Delta$: TouchAtlas $-$ VI smoother over the same presses (negative is better).
Bold: lowest, and significantly below the VI smoother.}
\label{tab:map}
\end{table}

We first show the tactile maps TouchAtlas builds and measure how accurately they place touch in the world map,
on surfaces the camera saw and on surfaces it did not (\cref{sec:maps}).
We then evaluate whether touch keeps contacts registered when the visual hand pose is unavailable,
and whether the gain comes from touch rather than from smoothing (\cref{sec:occlusion}).
Finally, we ablate what the surface constraints add beyond holding a contact in place (\cref{sec:ablations}).

\subsection{Setup}
\label{sec:setup}

\paragraph{Recordings.}
We record \ph{27} sequences of two people, with \ph{148}k RGB-D frames, \ph{296}k tactile frames and more than \ph{2{,}100} contact events.
The second person was recorded after all parameters were fixed.
Quantitative results come from a desk with everyday objects and a cylinder fixed to it.
A chair, an office setup and a round table are shown qualitatively.
The sequences contain contacts made while the back of the glove faces the camera, contacts under real occlusion,
and contacts on surfaces the head camera never sees.
Thresholds and noise parameters were set on development recordings (\cref{sec:app_setup}).

\paragraph{Ground truth.}
We use three independent sources.
\emph{Hand}: AprilTags~\cite{wang2016apriltag} on the back of the glove give the hand position, seen by the head camera and, under occlusion, by a static second camera.
They are masked out of the images the methods use.
\emph{Targets}: marks at known positions on the touched surfaces.
\emph{Surfaces}: a separate scan of each workspace that includes the faces hidden during the sequences.
Each part of it is labelled \emph{seen} or \emph{unseen} by whether the head camera observed it during a sequence.

\paragraph{Settings and metrics.}
\emph{Withheld visual pose}: where the head camera sees the glove tag, we remove all visual hand observations around every other contact,
from 0.3\,s before it to 0.3\,s after. On dense pressing, we also widen the window from 0.6 to \ph{\resWithheldWindowMax}\,s.
\emph{Real occlusion}: the hand is hidden behind an object, and the ground-truth hand position comes from the second camera.
We report the hand position error against the tag, in 3D and along and across the normal of the touched surface,
and the distance of a registered contact to its target and to the scanned surface.
All are medians over contact events.

\paragraph{Baselines.}
The \emph{VI smoother} is the estimator of \cref{sec:method} with all touch constraints removed,
so that any gain over it is due to touch rather than to smoothing.
The \emph{VI filter} is the causal fusion of \cref{sec:system}.
In both, the IMU contributes orientation only.
\emph{VI + map projection} projects the contacts of the VI smoother onto the visual map. It uses the map but leaves the trajectory unchanged.
HaWoR~\cite{zhang2025hawor} reconstructs world-space hand motion from the RGB video. We give it our glove detections and our camera poses.
We also run each touch constraint alone.
The no-slip constraint alone corresponds to the stance constraints used for feet and legs~\cite{foxlin2005pedestrian,hartley2020contact}.
\Cref{sec:app_setup} gives the calibration, the ground truth and the metrics in full.

\subsection{Tactile maps}
\label{sec:maps}
\label{sec:results}

\cref{fig:results_grid} shows three maps as they are built: a desk, a chair and a tag board with known targets.
With all vision available, contacts lie \ph{\resContactErr}\,mm from their targets (\cref{tab:map}).
Without a visual hand pose, TouchAtlas reduces the distance of contacts to the scan from \ph{\resSeenWithheldVI} to \ph{\resSeenWithheldTA}\,mm on surfaces the camera has seen,
and from \ph{\resUnseenWithheldVI} to \ph{\resUnseenWithheldTA}\,mm on surfaces it has not, relative to the VI smoother (\cref{tab:map}).
The hand itself is located to about 2\,cm, yet these contacts lie within a few millimetres of the surface,
at the resolution of the visual map (3\,mm voxels).
The surface constraint passes the accuracy of the map on to the contact.
Projecting the contacts of the VI smoother onto the visual map comes close in this setting (\ph{\resProjSeen} and \ph{\resProjUnseen}\,mm), but not behind a real occluder (\cref{sec:occlusion}).
For the second person, contacts lie \ph{\resWtwoTargetAll}\,mm from their targets with all vision.
With the visual pose withheld, they lie \ph{\resWtwoTargetWithheldTA}\,mm from them, against \ph{\resWtwoTargetWithheldVI}\,mm for the VI smoother (\cref{tab:wearer2}).
\subsection{Registration under occlusion}
\label{sec:occlusion}

\begin{table}[t]
\centering
\footnotesize\setlength{\tabcolsep}{2.5pt}\resfit{\columnwidth}{\begin{tabular}{@{}lcccc@{}}
\toprule
Median & \multicolumn{2}{c}{Visual pose withheld} & \multicolumn{2}{c}{Real occlusion} \\
\cmidrule(lr){2-3}\cmidrule(l){4-5}
{\scriptsize\color{black!55}[95\% interval]}, mm & Contact to target & Hand, 3D & Contact to scan & Hand, 3D \\
\midrule
VI filter & 36.1\ci{27.8}{44.0} & 31.6\ci{28.4}{35.5} & 24.9\ci{16.6}{36.9} & 85.8\ci{71.5}{96.3} \\
VI smoother & 33.6\ci{26.7}{40.6} & 27.4\ci{24.0}{30.4} & 51.1\ci{25.7}{55.7} & 149\ci{114}{226} \\
HaWoR & -- & 37.8\ci{32.2}{44.7} & -- & 238\ci{195}{293} \\
\rowcolor{oursrow}TouchAtlas (ours) & \textbf{18.9}\ci{15.5}{23.8} & \textbf{23.0}\ci{21.3}{26.2} & \textbf{14.4}\ci{6.4}{20.8} & 86.5\ci{64.0}{120} \\
\midrule
$\Delta$ (paired) & $-$14.7\ci{$-$21.3}{$-$7.3} & $-$4.4\ci{$-$7.5}{$-$0.3} & $-$36.7\ci{$-$48.1}{$-$13.4} & $-$62.4\ci{$-$137}{$-$20.6} \\
\bottomrule
\end{tabular}
}
\caption{\textbf{Registration without a visual hand pose.}
\emph{Visual pose withheld} around every other contact (\cref{tab:withheld_full}).
\emph{Real occlusion}: the hand behind an object (\cref{tab:real_occlusion}). Contact to scan is for the contacts behind the occluder.
\emph{Hand}: position error in contact.
$\Delta$: TouchAtlas $-$ VI smoother over the same events (negative is better).
Bold: lowest, and significantly below the VI smoother.}
\label{tab:withheld}
\end{table}

\begin{figure}[t]
\centering
\includegraphics[width=\columnwidth]{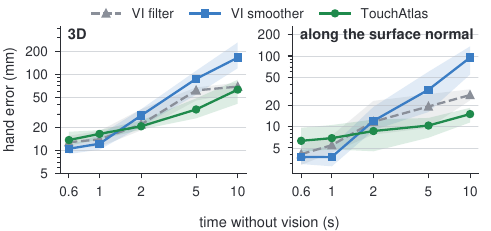}
\caption{\textbf{Error against the time without vision.}
Hand position error in contact (median and 95\% bootstrap band) with the visual hand pose withheld
for 0.6 to \resWithheldWindowMax\,s around a contact. Dense pressing on targets, two sequences (\cref{tab:window}).}
\label{fig:res_window}
\end{figure}

\begin{figure*}[t]
\centering
\includegraphics[width=\textwidth]{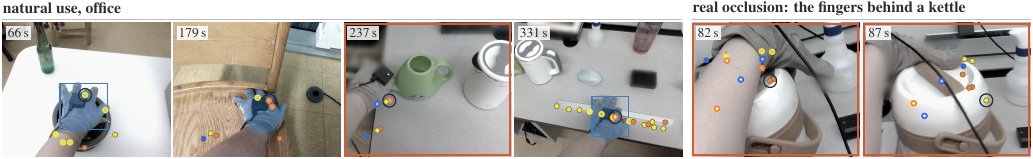}
\caption{\textbf{Contacts registered without a visual hand pose.}
Head-camera frames with the contacts registered so far, projected from their world positions and coloured by finger
(index: orange, middle: yellow, ring: blue). Hollow dots are hidden from the camera in that frame, and the current press is circled.
The visual hand pose is drawn in blue. Red frames have none.}
\label{fig:res_frames}
\end{figure*}

Vision and touch fail in complementary conditions. The hand is hidden exactly when it reaches behind or grasps an object, which is when it touches it.
A learned motion prior can only extrapolate through this interval, while a contact measures the hand in it.
With the visual pose withheld around a contact, the error of a contact on a known target drops from \ph{\resWithheldContactVI}\,mm for the VI smoother to \ph{\resWithheldContactTA}\,mm.
The hand position error is \ph{\resWithheldTA}\,mm, against \ph{\resWithheldVI}\,mm for the VI smoother and \ph{\resWithheldHawor}\,mm for HaWoR (\cref{tab:withheld}).
The gain of touch grows with the time without vision (\cref{fig:res_window}).
The error is \ph{\resSweepTwoTA} against \ph{\resSweepTwoVI}\,mm at 2\,s, \ph{\resSweepFiveTA} against \ph{\resSweepFiveVI}\,mm at 5\,s and \ph{\resSweepTenTA} against \ph{\resSweepTenVI}\,mm at 10\,s.
Along the surface normal, the error of TouchAtlas stays between \ph{\resAlongShortTA} and \ph{\resAlongLongTA}\,mm from \resSweepShortS{} to \resWithheldWindowMax\,s without vision,
while that of the VI smoother grows from \ph{\resAlongShortVI} to \ph{\resAlongLongVI}\,mm.
Each contact on a mapped surface is an absolute position fix, so the error no longer accumulates with time.
Behind a real occluder, TouchAtlas reaches \ph{\resRealOccTA}\,mm against \ph{\resRealOccVI}\,mm for the VI smoother and \ph{\resRealOccHawor}\,mm for HaWoR,
which infers depth from the apparent size of the partly visible hand (\cref{tab:withheld,fig:res_frames}).
Behind the occluder the hand error of TouchAtlas equals that of holding the last visual position,
but its contacts lie \ph{\resRealOccScanTA}\,mm from the scanned surface, against \ph{\resRealOccScanVIfilter}\,mm and \ph{\resRealOccScanVI}\,mm for the VI filter and the VI smoother.
Projection onto the visual map does not help here (\ph{\resProjRealOcc}\,mm), because the camera has not mapped the surfaces behind the occluder.

\subsection{Ablations}
\label{sec:ablations}

\begin{table}[t]
\centering
\footnotesize\setlength{\tabcolsep}{2.5pt}\resfit{\columnwidth}{\begin{tabular}{@{}lccc@{}}
\toprule
Median & \multirow{2}{*}{$k = 1$} & \multirow{2}{*}{$k = 2$} & \multirow{2}{*}{$k = 3$} \\
{\scriptsize\color{black!55}[95\% interval]}, mm &  &  &  \\
\midrule
VI smoother & 27.2\ci{22.4}{32.4} & 44.6\ci{30.9}{58.4} & 28.6\ci{24.1}{35.9} \\
\rowcolor{oursrow}TouchAtlas (ours) & 25.8\ci{22.4}{30.3} & \textbf{23.7}\ci{18.1}{30.2} & 25.2\ci{18.4}{32.7} \\
\midrule
$\Delta$ (paired) & $-$1.4\ci{$-$6.1}{$+$4.4} & $-$20.9\ci{$-$37.8}{$-$7.0} & $-$3.4\ci{$-$13.5}{$+$4.2} \\
\bottomrule
\end{tabular}
}
\caption{\textbf{Hand error by number of independent surface normals.}
Hand position error in contact with the visual pose withheld, for contacts on $k$ independent normals.
$\Delta$: TouchAtlas $-$ VI smoother over the same events (negative is better).
Bold: lowest, and significantly below the VI smoother.
Errors along and across the normal: \cref{tab:k_full}.}
\label{tab:k}
\end{table}

We withhold vision for 5\,s on dense pressing on targets (two sequences, one of them also used during development).
The surface constraint alone halves the error along the surface normal,
from \ph{\resAblFiveVIAlong} to \ph{\resAblFiveSurfAlong}\,mm, but leaves the 3D error at \ph{\resAblFiveSurfThreeD}\,mm against \ph{\resAblFiveVIThreeD}\,mm.
A surface fixes the hand only along its normal (\cref{tab:window}).
The no-slip constraint alone brings the 3D error to \ph{\resAblFiveNoSlipThreeD}\,mm.
Both together give \ph{\resAblFiveTAThreeD}\,mm in 3D, \ph{\resSweepFiveReduction\%} below the VI smoother, and \ph{\resAblFiveTAAlong}\,mm along the normal.
The two constraints remove different errors. Holding a contact stops the drift, and the surface removes the offset along its normal.
Neither alone reaches the accuracy of both.
With contacts on surfaces of one, two and three independent normals, TouchAtlas has the lower hand error in every case (\cref{tab:k}).
The difference is largest for two.
For \cref{prop:touch}\,(iii), letting patches built from contacts without vision constrain the hand raises its error from \ph{\resWeakTA} to \ph{\resWeakOK}\,mm
and brings the contacts no closer to the scan.
A touch surface is only as accurate as the hand poses that created it.
A patch built without vision carries their drift and feeds it back into the estimate.
The window length matters only for long occlusions (\cref{tab:window}).
Replayed in real time, the estimator runs at roughly 25\,Hz (\cref{tab:detection}).

\section{Conclusion}
\label{sec:conclusion}

We presented TouchAtlas, which builds world-registered tactile maps from tactile gloves and egocentric RGB-D cameras.
Touch has two roles in these maps.
Each contact is registered on the surface it touched,
and the contacts complete the camera's map with the surfaces only the hand has reached.
Both rely on the observation that the dependence between touch and hand pose runs both ways.
The hand pose places a contact in the world, and the contact constrains the hand pose.
A contact lies on the touched surface and, while it is held, stays at one point.
An online estimator combines these constraints with inertial orientation and intermittent vision,
and a contact on a mapped surface is registered on it without vision.
When the visual hand pose is unavailable, TouchAtlas reduces the median distance of contacts to a separate scan of the scene
from \ph{\resSeenWithheldVI} to \ph{\resSeenWithheldTA}\,mm ($-$\ph{78}\%) on surfaces the camera has seen,
and from \ph{\resUnseenWithheldVI} to \ph{\resUnseenWithheldTA}\,mm ($-$\ph{41}\%) on surfaces it has not, relative to visual--inertial tracking.
The error of a contact on a known target drops from \ph{\resWithheldContactVI} to \ph{\resWithheldContactTA}\,mm ($-$\ph{44}\%).

\paragraph{Limitations and future work.}
TouchAtlas assumes that the touched surfaces are static.
On a single surface, touch fixes the hand along the surface normal only.
Along the surface, and between contacts, the position follows the motion prior.
A surface the camera has not seen needs contacts made while the back of the hand is in view,
and one or two contacts recover its position but not its orientation.
Future work will let the touched objects move.

{
    \small
    \bibliographystyle{ieeenat_fullname}
    \bibliography{main}

\begin{thebibliography}{41}
\providecommand{\natexlab}[1]{#1}
\providecommand{\url}[1]{\texttt{#1}}
\expandafter\ifx\csname urlstyle\endcsname\relax
  \providecommand{\doi}[1]{doi: #1}\else
  \providecommand{\doi}{doi: \begingroup \urlstyle{rm}\Url}\fi

\bibitem[Bloesch et~al.(2012)Bloesch, Hutter, Hoepflinger, Leutenegger,
  Gehring, Remy, and Siegwart]{bloesch2012legged}
Michael Bloesch, Marco Hutter, Mark~A. Hoepflinger, Stefan Leutenegger,
  Christian Gehring, C.~David Remy, and Roland Siegwart.
\newblock State estimation for legged robots: Consistent fusion of leg
  kinematics and {IMU}.
\newblock In \emph{Robotics: Science and Systems (RSS)}, 2012.

\bibitem[Brahmbhatt et~al.(2019)Brahmbhatt, Ham, Kemp, and
  Hays]{brahmbhatt2019contactdb}
Samarth Brahmbhatt, Cusuh Ham, Charles~C. Kemp, and James Hays.
\newblock {ContactDB}: Analyzing and predicting grasp contact via thermal
  imaging.
\newblock In \emph{Proceedings of the IEEE/CVF Conference on Computer Vision
  and Pattern Recognition (CVPR)}, 2019.

\bibitem[Brahmbhatt et~al.(2020)Brahmbhatt, Tang, Twigg, Kemp, and
  Hays]{brahmbhatt2020contactpose}
Samarth Brahmbhatt, Chengcheng Tang, Christopher~D. Twigg, Charles~C. Kemp, and
  James Hays.
\newblock {ContactPose}: A dataset of grasps with object contact and hand pose.
\newblock In \emph{European Conference on Computer Vision (ECCV)}, 2020.

\bibitem[Campos et~al.(2021)Campos, Elvira, G{\'o}mez~Rodr{\'\i}guez, Montiel,
  and Tard{\'o}s]{campos2021orbslam3}
Carlos Campos, Richard Elvira, Juan~J. G{\'o}mez~Rodr{\'\i}guez, Jos{\'e} M.~M.
  Montiel, and Juan~D. Tard{\'o}s.
\newblock {ORB-SLAM3}: An accurate open-source library for visual,
  visual--inertial, and multimap {SLAM}.
\newblock \emph{IEEE Transactions on Robotics}, 37\penalty0 (6):\penalty0
  1874--1890, 2021.

\bibitem[Choi et~al.(2026)Choi, Park, Ji, Lee, Lee, Lee, and
  Lee]{choi2026robusthand}
Hyelim Choi, Hyunreal Park, Harim Ji, Somang Lee, Youngseon Lee, Yongseok Lee,
  and Dongjun Lee.
\newblock Robust hand tracking from visual-inertial fusion.
\newblock In \emph{2026 IEEE International Conference on Robotics and
  Automation (ICRA)}, 2026.

\bibitem[Dou et~al.(2024)Dou, Yang, Liu, Loquercio, and Owens]{dou2024tarf}
Yiming Dou, Fengyu Yang, Yi Liu, Antonio Loquercio, and Andrew Owens.
\newblock Tactile-augmented radiance fields.
\newblock In \emph{Proceedings of the IEEE/CVF Conference on Computer Vision
  and Pattern Recognition (CVPR)}, pages 26529--26539, 2024.

\bibitem[Foxlin(2005)]{foxlin2005pedestrian}
Eric Foxlin.
\newblock Pedestrian tracking with shoe-mounted inertial sensors.
\newblock \emph{IEEE Computer Graphics and Applications}, 25\penalty0
  (6):\penalty0 38--46, 2005.

\bibitem[Grady et~al.(2022)Grady, Tang, Brahmbhatt, Twigg, Wan, Hays, and
  Kemp]{grady2022pressurevision}
Patrick Grady, Chengcheng Tang, Samarth Brahmbhatt, Christopher~D. Twigg,
  Chengde Wan, James Hays, and Charles~C. Kemp.
\newblock {PressureVision}: Estimating hand pressure from a single {RGB} image.
\newblock In \emph{European Conference on Computer Vision (ECCV)}, 2022.

\bibitem[Guzov et~al.(2021)Guzov, Mir, Sattler, and Pons-Moll]{guzov2021hps}
Vladimir Guzov, Aymen Mir, Torsten Sattler, and Gerard Pons-Moll.
\newblock Human {POSEitioning} system ({HPS}): {3D} human pose estimation and
  self-localization in large scenes from body-mounted sensors.
\newblock In \emph{Proceedings of the IEEE/CVF Conference on Computer Vision
  and Pattern Recognition (CVPR)}, pages 4316--4327, 2021.

\bibitem[Hartley et~al.(2020)Hartley, Ghaffari, Eustice, and
  Grizzle]{hartley2020contact}
Ross Hartley, Maani Ghaffari, Ryan~M. Eustice, and Jessy~W. Grizzle.
\newblock Contact-aided invariant extended {Kalman} filtering for robot state
  estimation.
\newblock \emph{The International Journal of Robotics Research}, 39\penalty0
  (4):\penalty0 402--430, 2020.

\bibitem[Huang et~al.(2026)Huang, Mirzaee, Kaess, and Yuan]{huang2026gelslam}
Hung-Jui Huang, Mohammad~Amin Mirzaee, Michael Kaess, and Wenzhen Yuan.
\newblock {GelSLAM}: A real-time, high-fidelity, and robust {3D} tactile {SLAM}
  system.
\newblock \emph{IEEE Transactions on Robotics}, 2026.

\bibitem[Jiang et~al.(2024)Jiang, Xu, Li, Yu, Wang, Hu, Xie, Liu, Yang, Wang,
  Du, Tang, Zheng, Yao, Lu, and Liu]{jiang2024vitam}
Chunpeng Jiang, Wenqiang Xu, Yutong Li, Zhenjun Yu, Longchun Wang, Xiaotong Hu,
  Zhengyi Xie, Qingkun Liu, Bin Yang, Xiaolin Wang, Wenxin Du, Tutian Tang,
  Dongzhe Zheng, Siqiong Yao, Cewu Lu, and Jingquan Liu.
\newblock Capturing forceful interaction with deformable objects using a deep
  learning-powered stretchable tactile array.
\newblock \emph{Nature Communications}, 15:\penalty0 9513, 2024.

\bibitem[Jocher et~al.(2026)Jocher, Qiu, Liu, Lyu, Akyon, and
  Kalfaoglu]{jocher2026yolo26}
Glenn Jocher, Jing Qiu, Mengyu Liu, Shuai Lyu, Fatih~Cagatay Akyon, and
  Muhammet~Esat Kalfaoglu.
\newblock Ultralytics {YOLO26}: Unified real-time end-to-end vision models.
\newblock arXiv preprint arXiv:2606.03748, 2026.

\bibitem[Kou et~al.(2026)Kou, Kumar, Huang, Niehues, Mehta, Ristani, and
  Guan]{kou2026avihit}
Ziyi Kou, Ankit Kumar, Mia Huang, Taylor Niehues, Vatsal Mehta, Ergys Ristani,
  and Li Guan.
\newblock {AVI-HT}: Adaptive vision-{IMU} fusion for {3D} hand tracking.
\newblock arXiv preprint arXiv:2605.21714, 2026.

\bibitem[Ktistakis et~al.(2026)Ktistakis, Wang, Grande, and
  Sax]{ktistakis2026touchmapor}
Sophokles Ktistakis, Rui Wang, Bastian Grande, and Hugo Sax.
\newblock {TouchMap-OR}: Multi-view {3D} mapping of hand-surface contacts.
\newblock arXiv preprint arXiv:2605.17638, 2026.

\bibitem[Li et~al.(2026)Li, Wang, Yang, Zhu, Fang, Zeng, Han, Wang, Wang,
  Huang, Wang, Shimada, Luo, Li, Lv, Mukadam, Ranjan, Zhang, He, Liu, Chen,
  Pavone, Liu, Li, Tomizuka, and Fan]{li2026touchscale}
Dayou Li, Hao Wang, Qianqian Yang, Zihao Zhu, Haoquan Fang, Ziyao Zeng, Yan
  Han, Zihan Wang, Yan Wang, Baoru Huang, Dilin Wang, Kenji Shimada, Yiyue Luo,
  Manling Li, Teresa Lv, Mustafa Mukadam, Rakesh Ranjan, Ruohan Zhang, Qi He,
  Changliu Liu, Xu Chen, Marco Pavone, Bangya Liu, Jiachen Li, Masayoshi
  Tomizuka, and Zhiwen Fan.
\newblock {TouchScale}: 500 hours of human vision and touch for visual-tactile
  learning.
\newblock arXiv preprint arXiv:2610.10288, 2026.

\bibitem[Lin and Zhao(2026)]{lin2026artglove}
Changyi Lin and Ding Zhao.
\newblock {ART-Glove}: Articulated tactile glove for contact-grounded dexterous
  interaction capture.
\newblock arXiv preprint arXiv:2606.16370, 2026.

\bibitem[Lin et~al.(2026)Lin, Xia, Lai, Sun, Yang, and Pei]{lin2026imuhoi}
Lizhou Lin, Songpengcheng Xia, Zengyuan Lai, Lan Sun, Jiarui Yang, and Ling
  Pei.
\newblock {IMU-HOI}: A symbiotic framework for coherent human-object
  interaction and motion capture via contact-conscious inertial fusion.
\newblock In \emph{Proceedings of the IEEE/CVF Conference on Computer Vision
  and Pattern Recognition (CVPR)}, pages 42901--42910, 2026.

\bibitem[Liu et~al.(2017)Liu, Xie, Millar, Edmonds, Gao, Zhu, Santos, Rothrock,
  and Zhu]{liu2017poseforceglove}
Hangxin Liu, Xu Xie, Matt Millar, Mark Edmonds, Feng Gao, Yixin Zhu,
  Veronica~J. Santos, Brandon Rothrock, and Song-Chun Zhu.
\newblock A glove-based system for studying hand-object manipulation via joint
  pose and force sensing.
\newblock In \emph{2017 IEEE/RSJ International Conference on Intelligent Robots
  and Systems (IROS)}, pages 6617--6624, 2017.

\bibitem[Ni et~al.(2026)Ni, Zhang, Wei, Chen, Zhang, Shi, and
  Wang]{ni2026tactidex}
Suting Ni, Hanbing Zhang, Zhenyu Wei, Guo Chen, Chixuan Zhang, Ye Shi, and
  Jingya Wang.
\newblock {TactiDex}: A real-world tactile-guided benchmark for human-like
  dexterous manipulation.
\newblock arXiv preprint arXiv:2607.09190, 2026.

\bibitem[Pavlakos et~al.(2024)Pavlakos, Shan, Radosavovic, Kanazawa, Fouhey,
  and Malik]{pavlakos2024reconstructing}
Georgios Pavlakos, Dandan Shan, Ilija Radosavovic, Angjoo Kanazawa, David
  Fouhey, and Jitendra Malik.
\newblock Reconstructing hands in {3D} with transformers.
\newblock In \emph{Proceedings of the IEEE/CVF Conference on Computer Vision
  and Pattern Recognition (CVPR)}, 2024.

\bibitem[Qin et~al.(2026)Qin, Chen, Lin, Liu, Lyu, Su, Jin, Yu, Luo, Ding,
  Chen, and Xu]{qin2026dextouchwm}
Yan Qin, Yue Chen, Wenwei Lin, Shujia Liu, Chuqiao Lyu, Kailun Su, Weiyang Jin,
  Chenze Yu, Ping Luo, Wenbo Ding, Tianxing Chen, and Renjing Xu.
\newblock {DexTouch-WM}: Learning action-conditioned tactile world models from
  human touch for dexterous robot manipulation.
\newblock arXiv preprint arXiv:2609.20649 (IROS 2026 Workshop RoBoWoMo), 2026.

\bibitem[Ruppel and Zhang(2024)]{ruppel2024elasticglove}
Philipp Ruppel and Jianwei Zhang.
\newblock Elastic tactile sensor glove for dexterous teaching by demonstration.
\newblock \emph{Sensors}, 24\penalty0 (6):\penalty0 1912, 2024.

\bibitem[Shin et~al.(2026)Shin, Liu, and Molybog]{shin2026headcamera}
Yosub Shin, Yuan Liu, and Igor Molybog.
\newblock A head camera is (almost) all you need: An electronics-free {UMI}
  glove labelled and observed by one moving camera.
\newblock arXiv preprint arXiv:2610.27622, 2026.

\bibitem[Song et~al.(2025)Song, Li, Fu, Murphy, Zhou, Shiv, Li, Xiong, Owens,
  Du, Luo, Cheng, Torralba, Matusik, and Liang]{song2025opentouch}
Yuxin~Ray Song, Jinzhou Li, Rao Fu, Devin Murphy, Kaichen Zhou, Rishi Shiv,
  Yaqi Li, Haoyu Xiong, Crystal~Elaine Owens, Yilun Du, Yiyue Luo, Xianyi
  Cheng, Antonio Torralba, Wojciech Matusik, and Paul~Pu Liang.
\newblock {OpenTouch}: Bringing full-hand touch to real-world interaction.
\newblock arXiv preprint arXiv:2512.16842, 2025.

\bibitem[Sundaram et~al.(2019)Sundaram, Kellnhofer, Li, Zhu, Torralba, and
  Matusik]{sundaram2019stag}
Subramanian Sundaram, Petr Kellnhofer, Yunzhu Li, Jun-Yan Zhu, Antonio
  Torralba, and Wojciech Matusik.
\newblock Learning the signatures of the human grasp using a scalable tactile
  glove.
\newblock \emph{Nature}, 569\penalty0 (7758):\penalty0 698--702, 2019.

\bibitem[Suresh et~al.(2021)Suresh, Bauza, Yu, Mangelson, Rodriguez, and
  Kaess]{suresh2021tactileslam}
Sudharshan Suresh, Maria Bauza, Kuan-Ting Yu, Joshua~G. Mangelson, Alberto
  Rodriguez, and Michael Kaess.
\newblock Tactile {SLAM}: Real-time inference of shape and pose from planar
  pushing.
\newblock In \emph{2021 IEEE International Conference on Robotics and
  Automation (ICRA)}, pages 11322--11328, 2021.

\bibitem[Suresh et~al.(2024)Suresh, Qi, Wu, Fan, Pineda, Lambeta, Malik,
  Kalakrishnan, Calandra, Kaess, Ortiz, and Mukadam]{suresh2024neuralfeels}
Sudharshan Suresh, Haozhi Qi, Tingfan Wu, Taosha Fan, Luis Pineda, Mike
  Lambeta, Jitendra Malik, Mrinal Kalakrishnan, Roberto Calandra, Michael
  Kaess, Joseph Ortiz, and Mustafa Mukadam.
\newblock Neural feels with neural fields: Visuo-tactile perception for in-hand
  manipulation.
\newblock \emph{Science Robotics}, page adl0628, 2024.

\bibitem[Swann et~al.(2024)Swann, Strong, Do, Camps, Schwager, and
  Kennedy]{swann2024touchgs}
Aiden Swann, Matthew Strong, Won~Kyung Do, Gadiel~Sznaier Camps, Mac Schwager,
  and Monroe Kennedy, III.
\newblock {Touch-GS}: Visual-tactile supervised {3D} {G}aussian splatting.
\newblock In \emph{2024 IEEE/RSJ International Conference on Intelligent Robots
  and Systems (IROS)}, pages 10511--10518, 2024.

\bibitem[Wang and Olson(2016)]{wang2016apriltag}
John Wang and Edwin Olson.
\newblock {AprilTag 2}: Efficient and robust fiducial detection.
\newblock In \emph{2016 IEEE/RSJ International Conference on Intelligent Robots
  and Systems (IROS)}, pages 4193--4198, 2016.

\bibitem[Yang et~al.(2026)Yang, Li, Liu, Zhu, Chen, Xu, Hu, Yu, Friesen, Zheng,
  Hipwell, Qian, Shimada, and Fan]{yang2026ego2contact}
Qianqian Yang, Dayou Li, Bangya Liu, Zihao Zhu, Nuo Chen, Siyuan Xu, Hezhen Hu,
  Chen Yu, Rebecca Friesen, Minghui Zheng, Cynthia Hipwell, Xiaoning Qian,
  Kenji Shimada, and Zhiwen Fan.
\newblock {Ego2Contact}: Predicting touch from egocentric human video for
  contact-rich robot manipulation.
\newblock arXiv preprint arXiv:2610.22932, 2026.

\bibitem[Yi et~al.(2023)Yi, Zhou, Habermann, Golyanik, Pan, Theobalt, and
  Xu]{yi2023egolocate}
Xinyu Yi, Yuxiao Zhou, Marc Habermann, Vladislav Golyanik, Shaohua Pan,
  Christian Theobalt, and Feng Xu.
\newblock {EgoLocate}: Real-time motion capture, localization, and mapping with
  sparse body-mounted sensors.
\newblock \emph{ACM Transactions on Graphics}, 42\penalty0 (4), 2023.

\bibitem[Yu et~al.(2025{\natexlab{a}})Yu, Xu, Xie, Li, Anthony, Zhang, and
  Lu]{yu2025vitamd}
Zhenjun Yu, Wenqiang Xu, Pengfei Xie, Yutong Li, Brian~W. Anthony, Zhuorui
  Zhang, and Cewu Lu.
\newblock Dynamic reconstruction of hand-object interaction with distributed
  force-aware contact representation.
\newblock In \emph{Proceedings of the IEEE/CVF International Conference on
  Computer Vision (ICCV)}, pages 8590--8599, 2025{\natexlab{a}}.

\bibitem[Yu et~al.(2025{\natexlab{b}})Yu, Zafeiriou, and Birdal]{yu2025dynhamr}
Zhengdi Yu, Stefanos Zafeiriou, and Tolga Birdal.
\newblock {Dyn-HaMR}: Recovering {4D} interacting hand motion from a dynamic
  camera.
\newblock In \emph{Proceedings of the IEEE/CVF Conference on Computer Vision
  and Pattern Recognition (CVPR)}, pages 27716--27726, 2025{\natexlab{b}}.

\bibitem[Zhang et~al.(2024)Zhang, Ren, Yuan, Zhao, Li, Sun, Liang, Yu, Shen,
  and Cao]{zhang2024mmvp}
He Zhang, Shenghao Ren, Haolei Yuan, Jianhui Zhao, Fan Li, Shuangpeng Sun,
  Zhenghao Liang, Tao Yu, Qiu Shen, and Xun Cao.
\newblock {MMVP}: A multimodal {MoCap} dataset with vision and pressure
  sensors.
\newblock In \emph{Proceedings of the IEEE/CVF Conference on Computer Vision
  and Pattern Recognition (CVPR)}, 2024.

\bibitem[Zhang et~al.(2025)Zhang, Deng, Ma, and Potamias]{zhang2025hawor}
Jinglei Zhang, Jiankang Deng, Chao Ma, and Rolandos~Alexandros Potamias.
\newblock {HaWoR}: World-space hand motion reconstruction from egocentric
  videos.
\newblock In \emph{Proceedings of the IEEE/CVF Conference on Computer Vision
  and Pattern Recognition (CVPR)}, pages 1805--1815, 2025.

\bibitem[Zhang et~al.(2026{\natexlab{a}})Zhang, Yuan, Zhang, Cheng, and
  Gao]{zhang2026egotac}
Wenkang Zhang, Chengbo Yuan, Zicheng Zhang, Zhengxue Cheng, and Yang Gao.
\newblock {EgoTac}: In-the-wild tactile prediction from egocentric vision.
\newblock In \emph{Advances in Neural Information Processing Systems
  (NeurIPS)}, 2026{\natexlab{a}}.

\bibitem[Zhang et~al.(2026{\natexlab{b}})Zhang, Kou, Qin, Huang, Ristani,
  Kumar, Chen, He, Boularias, and Guan]{zhang2026glove2hand}
Xinyu Zhang, Ziyi Kou, Chuan Qin, Mia Huang, Ergys Ristani, Ankit Kumar, Lele
  Chen, Kun He, Abdeslam Boularias, and Li Guan.
\newblock {Glove2Hand}: Synthesizing natural hand-object interaction from
  multi-modal sensing gloves.
\newblock In \emph{Proceedings of the IEEE/CVF Conference on Computer Vision
  and Pattern Recognition (CVPR)}, 2026{\natexlab{b}}.

\bibitem[Zhao et~al.(2025)Zhao, Kwon, Streli, Pollefeys, and
  Holz]{zhao2025egopressure}
Yiming Zhao, Taein Kwon, Paul Streli, Marc Pollefeys, and Christian Holz.
\newblock {EgoPressure}: A dataset for hand pressure and pose estimation in
  egocentric vision.
\newblock In \emph{Proceedings of the IEEE/CVF Conference on Computer Vision
  and Pattern Recognition (CVPR)}, 2025.

\bibitem[Zhou et~al.(2026{\natexlab{a}})Zhou, Lu, Lin, Chen, Lyu, and
  Ding]{zhou2026touchsight}
Danyan Zhou, Jinxuan Lu, Jiawei Lin, Tianxing Chen, Chuqiao Lyu, and Wenbo
  Ding.
\newblock {TouchSight}: Bare-handed tactile prediction from egocentric video
  via generative visual augmentation.
\newblock arXiv preprint arXiv:2609.20414, 2026{\natexlab{a}}.

\bibitem[Zhou et~al.(2026{\natexlab{b}})Zhou, Gao, Hong, Liu, Zhang, Dai, Zhen,
  Lyu, Wu, Mao, Wang, Jiang, Ding, and Yang]{zhou2026touchanything}
Jianyi Zhou, Ziteng Gao, Feiyang Hong, Zirui Liu, Guannan Zhang, Weisheng Dai,
  Ruichen Zhen, Chuqiao Lyu, Haotian Wu, Yinian Mao, Xushi Wang, Yuxiang Jiang,
  Wenbo Ding, and Shuo Yang.
\newblock {TouchAnything}: A dataset and framework for bimanual tactile
  estimation from egocentric video.
\newblock arXiv preprint arXiv:2605.13083, 2026{\natexlab{b}}.

\end{thebibliography}
}

\clearpage
\setcounter{page}{1}
\maketitlesupplementary
\appendix
\renewcommand{\thetable}{S\arabic{table}}\setcounter{table}{0}
\renewcommand{\thefigure}{S\arabic{figure}}\setcounter{figure}{0}
\renewcommand{\theHtable}{S\arabic{table}}\renewcommand{\theHfigure}{S\arabic{figure}}
\makeatletter\let\Hy@backout\@gobble\makeatother
\label{supp:start}

\section{Experimental setup in full}
\label{sec:app_setup}

\paragraph{Participants.}
The two people who wear the glove are authors of this paper and gave their informed consent.
No ethics review was sought, since the recordings involve no participants other than the authors.
The recordings show only the gloved hand and forearm of the person wearing the glove.

\paragraph{Recordings.}
We record \ph{27} long sequences of two people, with \ph{148}k RGB-D frames, \ph{296}k tactile frames and more than \ph{2{,}100} contact events.
The workspace of the quantitative results is a desk with boxes, everyday objects and a cylindrical lunch box fixed to it.
These surfaces are static and their normals are independent.
Each sequence starts with a short sweep of the head camera over the scene without the hand, which builds an initial map of the scene.
The sweep covers the scene from the side of the person only.
The person then presses each surface repeatedly.
The sequences contain contacts made while the back of the glove faces the camera, contacts under real occlusion
(the hand behind the box, curled, or near the edge of the view),
and contacts on surfaces the head camera never sees during the sequence.
An example is the back of the box, where the camera sees only the back of the hand or does not see the hand at all.
We store all color and depth frames at 30\,Hz, together with all glove streams.
\looseness=-1

\begin{figure}[b]
\centering
\includegraphics[width=\columnwidth]{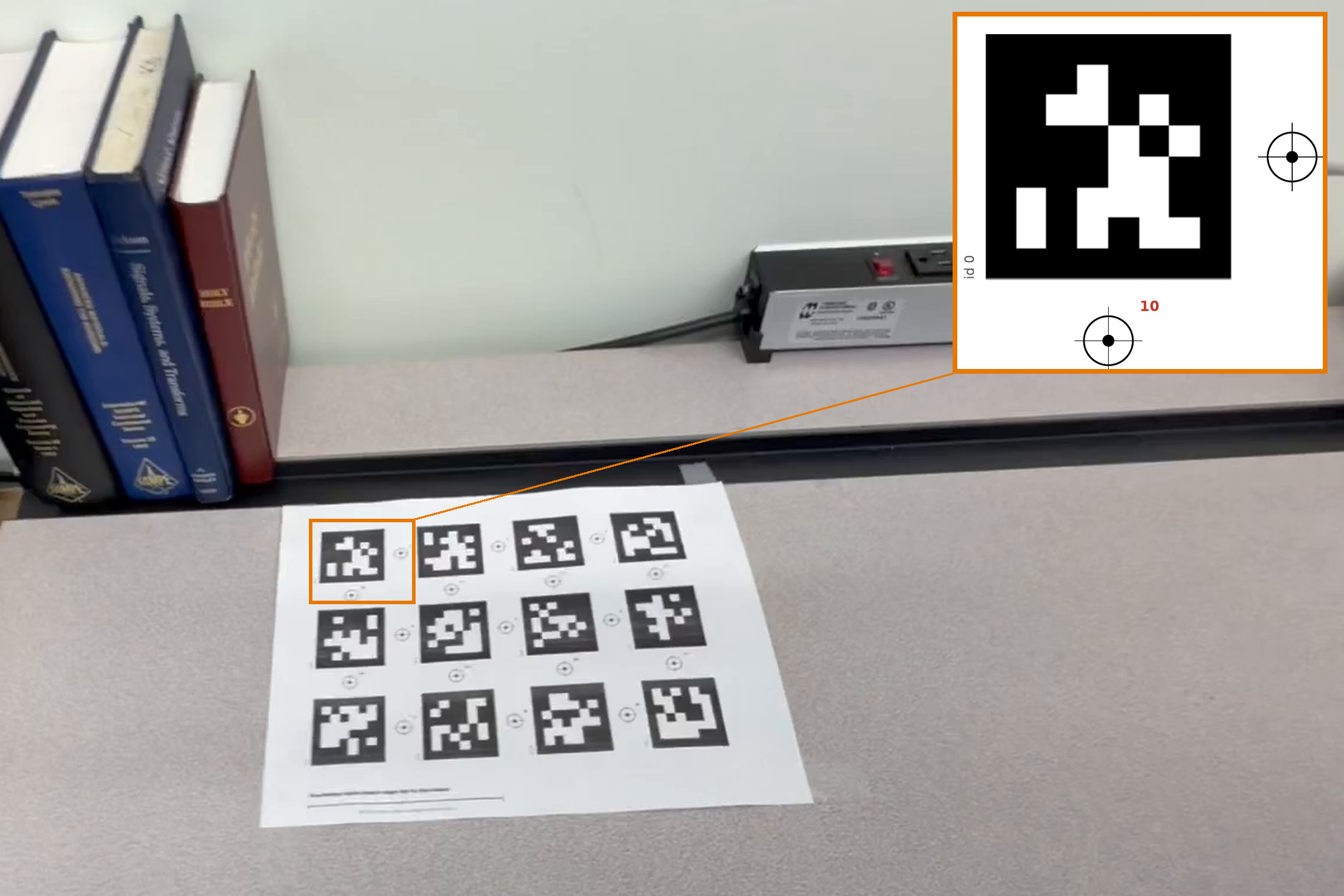}
\caption{\textbf{Tag board.} A printed sheet of AprilTags attached to the desk.
Small circles between the tags are the printed dots used as targets.}
\label{fig:supp_board}
\end{figure}

\paragraph{Calibration.}
Three quantities are calibrated.
The first is the depth of the camera, an Intel RealSense D435.
It is corrected by a scale factor for each recording (\ph{0.976--0.987}), measured on the tag board, a board of AprilTags of known size on the desk (\cref{fig:supp_board}).
The factor brings the depth to the metric scale of the board, so that the map and the hand positions share the scale of the ground truth.
It is applied to every method.
The second is the fingertip pad correction $\mathbf{c}_r$ of \cref{eq:pad}.
It is fitted on a separate sequence in which each finger presses \ph{17} printed dots.
It is the constant that places the pad on the dot under the ground-truth hand pose.
For example, for the index finger it moves the contact point by \ph{25}\,mm towards the palmar side.
If the glove is calibrated again, the new correction replaces the old one only when it is at least as accurate.
The accuracy is the error on each press when that press is left out of the fit.
The third is the offset from the glove tag to the hand position.
It is the median offset to the visual hand position of \cref{sec:system} over the frames of the calibration sequence in which both are valid.
It is fixed for all evaluated sequences.
\widowpenalty=10000\par\widowpenalty=150

\paragraph{Ground truth.}
We use three sources that do not depend on each other.
\begin{itemize}
\setlength{\itemsep}{2pt}\setlength{\parskip}{0pt}\setlength{\topsep}{2pt}
\item \emph{Hand.} AprilTags~\cite{wang2016apriltag} on the back of the glove give the hand position.
The tag board on the desk (\cref{fig:supp_board}) gives the camera pose.
The ORB-SLAM3 world frame is aligned to the board frame from the camera poses of the frames that see the tag board.
The glove tags are observed by the head camera and by a static second camera, a phone on a tripod placed to the side (\ph{1080p at 30\,Hz}).
The pose of the second camera is estimated from the tag board.
The second camera often sees the glove tag when the head camera does not, \eg when the hand is curled or behind the box.
Real occlusion therefore also has ground-truth hand positions.
The two cameras agree on the glove tag to a median of \ph{3--8}\,mm per sequence, perpendicular to the viewing ray of the second camera.
The glove tags are masked out of the head-camera images before hand detection, so that neither TouchAtlas nor the baselines use them.
The tag-to-hand offset is calibrated against the visual hand position.
A constant bias of the visual hand position is therefore not visible in this ground truth.
The next two sources do not share this limit.
\item \emph{Targets.} The targets are the dots printed on the tag board (\cref{fig:supp_board}), whose positions in the board frame are known.
A contact made on a target should be registered at the target.
This check is independent of the hand model and of the hand ground truth.
\item \emph{Surfaces.} After the sequences of a workspace, with the objects untouched, the head camera is moved around them without the hand.
\textbf{This scan includes the faces the camera did not see during the sequences.}
The depth is fused into a mesh in the board frame.
\end{itemize}

\paragraph{Evaluation settings.}
\begin{itemize}
\setlength{\itemsep}{2pt}\setlength{\parskip}{0pt}\setlength{\topsep}{2pt}
\item \emph{Withheld visual pose.} Where the head camera sees the glove tag, we remove all visual hand observations around every other contact, from 0.3\,s before it to 0.3\,s after.
The removed observations are the hand position and the HaMeR keypoints used by the finger EKF.
On dense pressing, we also widen the window from 0.6\,s to \ph{\resWithheldWindowMax}\,s.
The error is then reported as a function of how long the hand goes unobserved.
\item \emph{Real occlusion.} The hand is hidden behind an object.
In these segments the head camera does not give a reliable hand pose.
We compare our result with the hand position from the second camera, with the targets and with the scan.
A visual hand pose is unreliable when no glove is detected or the palm depth check of \cref{sec:system} fails.
\item \emph{Unseen surfaces.} We evaluate the contacts on surfaces that the head camera did not see during the sequence.
\end{itemize}

\paragraph{Metrics.}
\begin{itemize}
\setlength{\itemsep}{2pt}\setlength{\parskip}{0pt}\setlength{\topsep}{2pt}
\item \emph{Hand position error.} The error against the tag ground truth, in 3D and split into the components along and across the normal of the touched surface.
It is reported separately for frames in contact and frames not in contact, since touch only constrains the position during contact.
It isolates the estimator, because the hand model does not enter.
\item \emph{Contact-to-target error.} How far a registered contact lies from the target that was pressed.
\item \emph{Contact-to-surface error.} The distance between a registered contact and the scanned surface, split into seen and unseen surfaces.
\item \emph{Surface accuracy.} The distance and the normal angle between the touch patches and the scanned surface.
\item \emph{Latency.} How long it takes until the position of a contact no longer changes.
\end{itemize}

\paragraph{Baselines and ablations.}
The main baseline is the \emph{VI smoother}.
It is the estimator of \cref{sec:method} with all touch constraints removed, a visual--inertial smoother over the same window.
Any gain over it is therefore due to touch rather than to smoothing.
The \emph{VI filter} is the causal fusion of \cref{sec:system}.
In both cases, the IMU only contributes orientation.
\emph{VI + map projection} uses the map of the camera without touch constraints.
The contacts registered by the VI smoother are projected onto the visual map along its normal.
This is the best one can do with the map of the camera without changing the trajectory.
It cannot place a contact on a surface the camera has not seen, because that surface is not in the map.
HaWoR~\cite{zhang2025hawor} reconstructs world-space hand motion from the RGB video.
We give it our glove detections and our camera poses.
We then add the touch constraints one at a time.
The no-slip constraint alone is like the stance constraints used for feet and legs~\cite{foxlin2005pedestrian,hartley2020contact}.
The surface constraint alone and both together (TouchAtlas) show what the map adds.
We further vary the window length $L$.

\section{Implementation details}
\label{sec:app_impl}

\paragraph{Palm depth fit (\cref{sec:system}).}
The palm is fitted to the glove fabric in the depth image of the RealSense D435.
The translation along the viewing ray starts from the larger depth offsets, so that the fit lands on the fabric and not on what stands out from it.
It is then refined on the samples within 12\,mm of the model.
We accept the fit only if enough palm samples lie within 12\,mm of the fitted palm.
Enough means at least 25\% of the samples and no fewer than 12.
In addition, the reprojection and the surface normal of the fitted depth must be consistent with the HaMeR palm.
The median reprojection error must be below 12\,px and the normals within 35$^\circ$.

\paragraph{Real-time implementation.}
Camera reading, ORB-SLAM3 tracking, hand estimation and output encoding run in separate threads.
Camera tracking therefore keeps every frame, while the hand pipeline takes the latest one.
We optimize the HaMeR backbone for inference and vectorize the palm depth sampling.
This reduces hand estimation from 84 to 31\,ms per frame.
On an RTX 4070 Ti SUPER, camera tracking runs at 30\,Hz and hand estimation at 25\,Hz with the hand in view (\cref{tab:detection}).

\paragraph{Surfaces (\cref{sec:map}).}
\emph{Visual map.} The visual map is fused from the depth images of the head camera in a TSDF with 3\,mm voxels.
The camera sees only the faces of an object that are turned towards it, so the visual map ends at the edges of the object.
Near this boundary, the visual map returns no surface for a query.
The reason is that a contact made just behind an edge lies close to the visible face.
If the visual map answered, the contact would be pulled onto that face.

\emph{Touch surface.} Each patch of the touch surface is a disc with a radius of 2\,cm.
For a query, we blend the tangent planes of the patches near the query point.
A contact made while the hand is not seen has a less accurate position.
Such a contact joins an existing patch if it lies within 2\,cm of it along the patch, but it does not move the patch.
The normal of a patch is computed from its contacts when they spread over at least 4\,mm in two directions.
When they do not, the normal is the direction midway between the pad normal and the fingertip axis.

\paragraph{Derivation of \cref{prop:touch}.}
Each statement follows from the form of the residuals.
Where the surface is locally flat, the surface residual of \cref{eq:surf} depends on the hand position $\mathbf{p}_t$ only through $\mathbf{n}^\top \mathbf{p}_t$.
This gives (i).
The no-slip residual of \cref{eq:slip} depends on $\mathbf{p}_t$ only through $\mathbf{p}_t - \mathbf{a}_H$.
This gives (ii).
When the surface itself is unknown, both residuals depend only on differences between these quantities.
This gives (iii).

\begin{figure}[t]
\centering
\includegraphics[width=\columnwidth]{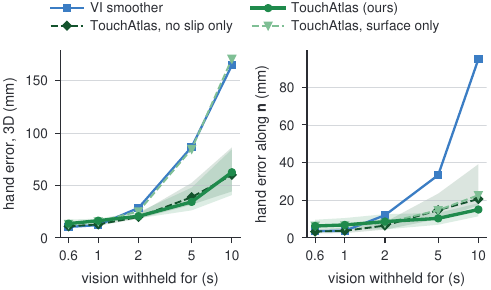}
\caption{\textbf{Which constraint gives which part of the gain.}
Hand position error during contact, in 3D (left) and along the surface normal (right).
The curves are the VI smoother, TouchAtlas, and TouchAtlas with each touch constraint alone.
No slip alone gives most of the gain in 3D, and the surface reduces the error along its normal.
With 10\,s without vision, TouchAtlas reduces the 3D error from 165 to 63\,mm and the error along the normal from 95 to 15\,mm.
The numbers are in \cref{tab:window}.}
\label{fig:res_ablation}
\end{figure}

\paragraph{Estimator (\cref{sec:estimator}).}
\Cref{eq:energy} has four terms and a prior.
We give their details here.
The first term keeps the hand position close to the visual hand position $\hat{\mathbf{p}}_t$.
It acts on the frames $\mathcal{V}$, in which the depth fit of \cref{sec:system} is valid.
The visual hand position is brought into the world frame with the camera pose from ORB-SLAM3.
Its standard deviation is $\sigma_V = 10$\,mm.
The second and fourth terms are the two touch constraints, the surface constraint and the no-slip constraint.
Their standard deviation is $\sigma_S = \sigma_H = 3$\,mm.
The third term favors a constant velocity.
Its standard deviation is $\sigma_{A,t} = \sigma_A\,\Delta t^2$ with $\sigma_A = 3$\,m/s$^2$.
When neither vision nor touch is available, it is the only term that acts on the position.
$\rho$ is a Huber loss.
It limits the effect of wrong detections, wrong surface associations and undetected slip.
When a position leaves the window, its information is kept in $E_\text{prior}$.

The surfaces enter through \cref{eq:surf} and are updated in alternation with the trajectory.
The visual map is fused causally.
Every depth frame is integrated as it arrives.
The mesh is extracted again once per second in a background thread.
A contact is therefore constrained only by surfaces the camera has already seen.
The touch surface is refitted whenever its contacts have moved by more than 1\,mm.
After each solve, the contact events in the window are associated again.
If an association changed, the window is solved once more.
A patch stays in the map after its contacts have left the window and keeps its place.
Touching it again therefore constrains the trajectory to the earlier, visually anchored contacts.

Each new frame adds its state and factors.
We then run damped Gauss--Newton iterations on \cref{eq:energy}.
An update takes \ph{5--12} iterations on average and starts from the result of the previous update.
The contacts in the window are registered again with \cref{eq:register}, using the refined positions.
A contact is final once its event has ended and left the window.
With $L = \ph{\resWindowL}$\,s, this is about \ph{\resWindowL}\,s after the hand lifts.
Replayed in real time, the estimator runs at roughly 25\,Hz (\cref{tab:detection}).

\section{Additional results}
\label{sec:app_results}

\makeatletter\setlength{\@dblfptop}{0pt}\setlength{\@dblfpsep}{16pt}\setlength{\@dblfpbot}{0pt plus 1fil}\makeatother

\par\vspace{6pt}\noindent\begin{minipage}{\columnwidth}
\centering
\small
\footnotesize\setlength{\tabcolsep}{2.5pt}\resfit{\columnwidth}{\begin{tabular}{@{}lcccc@{}}
\toprule
Hand & Face & Dist.\ p50 & Dist.\ p90 & Angle p50 \\
\midrule
\rowcolor{oursrow}observed & seen & 3.9--21.2 & 3.9--38.6 & 36.7--79.9$^\circ$ \\
\rowcolor{oursrow}observed & unseen & 5.4--13.5 & 6.2--34.8 & 29.9--66.5$^\circ$ \\
\addlinespace[3pt]
not observed & seen & 12.4--21.0 & 17.8--65.0 & 40.0--67.3$^\circ$ \\
not observed & unseen & 9.8--39.3 & 15.6--100 & 46.1--70.7$^\circ$ \\
\bottomrule
\end{tabular}
}
\captionof{table}{\textbf{Touch patches against the scan.}
How far the touch patches lie from the scan, and how much their normals differ.
\emph{Hand}: whether the head camera observed the hand when the patch was created.
\emph{Face}: whether the head camera saw the face.
Only patches created with the hand observed are used as constraints (shaded).
Distances are in mm.}
\label{tab:patches}

\vspace{12pt}

\centering
\footnotesize\setlength{\tabcolsep}{4pt}\resfit{\columnwidth}{\begin{tabular}{@{}lcc@{}}
\toprule
 & \multicolumn{2}{c}{Contact to target (mm)} \\
\cmidrule(l){2-3}
Method & all vision & visual pose withheld \\
\midrule
VI filter & 17.3\ci{15.2}{20.8} & 34.5\ci{29.1}{47.5} \\
VI smoother & 17.5\ci{14.3}{19.0} & 35.6\ci{28.4}{41.4} \\
\rowcolor{oursrow}TouchAtlas (ours) & \textbf{14.8}\ci{12.6}{15.9} & \textbf{25.1}\ci{21.4}{34.3} \\
\midrule
$\Delta$ (paired) & $-$2.7\ci{$-$5.5}{$-$0.1} & $-$10.6\ci{$-$17.1}{$-$1.0} \\
\bottomrule
\end{tabular}
}
\captionof{table}{\textbf{Second person}, recorded after all parameters were fixed.
Dense pressing on known targets.
\emph{Contact to target}: how far a registered contact lies from the target that was pressed, with all vision and with the visual pose withheld.
$\Delta$: TouchAtlas $-$ VI smoother (negative is better).
Bold: lowest.}
\label{tab:wearer2}
\end{minipage}

\begin{table*}[t]
\centering
\small
\setlength{\tabcolsep}{4pt}\resfit{\textwidth}{\begin{tabular}{@{}lcccccc@{}}
\toprule
 & \multicolumn{2}{c}{Contact to target} & \multicolumn{4}{c}{Contact to scanned surface} \\
\cmidrule(lr){2-3}\cmidrule(l){4-7}
Method & 3D & p90 & seen & unseen & seen, no visual pose & unseen, no visual pose \\
\midrule
VI filter & 16.7\ci{12.5}{20.9} & 35.2 & 17.6\ci{15.3}{19.0} & 15.2\ci{10.5}{17.4} & 18.1\ci{15.3}{19.6} & 15.1\ci{9.0}{20.8} \\
VI smoother & 16.0\ci{11.5}{21.6} & 32.1 & 16.0\ci{13.8}{18.9} & 16.6\ci{12.8}{19.3} & 16.2\ci{13.7}{20.5} & 11.4\ci{8.6}{17.4} \\
VI + map projection & 15.5\ci{13.8}{17.5} & \textbf{25.2} & 4.0\ci{3.1}{4.9} & \textbf{8.8}\ci{7.6}{12.8} & 5.2\ci{3.2}{7.8} & 7.5\ci{5.6}{14.0} \\
\rowcolor{oursrow}TouchAtlas (ours) & \textbf{14.6}\ci{13.0}{17.2} & 25.9 & \textbf{3.3}\ci{3.0}{3.9} & 9.1\ci{7.6}{12.7} & \textbf{3.6}\ci{3.0}{4.7} & \textbf{6.7}\ci{4.3}{9.2} \\
\midrule
$\Delta$ (paired) & $-$1.4\ci{$-$6.2}{$+$2.0} & -- & $-$12.7\ci{$-$15.5}{$-$10.5} & $-$7.5\ci{$-$10.2}{$-$3.3} & $-$12.5\ci{$-$16.8}{$-$10.0} & $-$4.7\ci{$-$11.3}{$-$1.9} \\
\bottomrule
\end{tabular}
}
\caption{\textbf{Map accuracy, all metrics.}
\emph{Contact to target}: the distance from a contact to the target that was pressed.
\emph{Contact to scanned surface}: the distance from a contact to the scan, on surfaces the head camera saw or did not see.
$\Delta$: TouchAtlas $-$ VI smoother (negative is better).
Values are in mm. Bold: lowest.}
\label{tab:map_full}
\end{table*}

\begin{table*}[t]
\centering
\small
\setlength{\tabcolsep}{4pt}\resfit{\textwidth}{\begin{tabular}{@{}lc>{\columncolor{oursrow}}cc@{\hspace{10pt}}c>{\columncolor{oursrow}}cc@{\hspace{10pt}}c>{\columncolor{oursrow}}cc@{}}
\toprule
 & \multicolumn{3}{c}{Hand position, 3D} & \multicolumn{3}{c}{Hand position at $t$, 3D} & \multicolumn{3}{c}{Contact to target} \\
\cmidrule(lr){2-4}\cmidrule(lr){5-7}\cmidrule(l){8-10}
Data & VI & \multicolumn{1}{c}{Ours} & $\Delta$ & VI & \multicolumn{1}{c}{Ours} & $\Delta$ & VI & \multicolumn{1}{c}{Ours} & $\Delta$ \\
\midrule
Objects & \textbf{21.7} & 25.1 & $+$3.4\ci{$-$1.2}{$+$7.5}\,{\scriptsize($+$15\%)} & 56.3 & \textbf{27.3} & $-$28.9\ci{$-$36.9}{$-$22.0}\,{\scriptsize($-$51\%)} & 36.1 & \textbf{17.0} & $-$19.1\ci{$-$31.2}{$-$11.7}\,{\scriptsize($-$53\%)} \\
Objects, coverage & 24.4 & \textbf{21.3} & $-$3.1\ci{$-$11.8}{$+$6.2}\,{\scriptsize($-$13\%)} & 43.8 & \textbf{27.4} & $-$16.3\ci{$-$38.0}{$-$7.0}\,{\scriptsize($-$37\%)} & -- & -- & -- \\
Targets & 38.5 & \textbf{22.6} & $-$15.9\ci{$-$30.1}{$-$7.0}\,{\scriptsize($-$41\%)} & 92.7 & \textbf{21.3} & $-$71.4\ci{$-$99.9}{$-$51.6}\,{\scriptsize($-$77\%)} & 33.1 & \textbf{19.9} & $-$13.1\ci{$-$28.3}{$-$4.4}\,{\scriptsize($-$40\%)} \\
\midrule
Pooled & 27.4 & \textbf{23.0} & $-$4.4\ci{$-$7.5}{$-$0.3}\,{\scriptsize($-$16\%)} & 67.0 & \textbf{24.0} & $-$43.0\ci{$-$53.0}{$-$33.7}\,{\scriptsize($-$64\%)} & 33.6 & \textbf{18.9} & $-$14.7\ci{$-$21.3}{$-$7.3}\,{\scriptsize($-$44\%)} \\
\bottomrule
\end{tabular}
}
\caption{\textbf{Registration with the visual hand pose withheld, by data group.}
\emph{VI}: the VI smoother. \emph{Ours}: TouchAtlas.
\emph{Hand position}: the median error over the frames in contact.
\emph{At $t$}: the estimate for the newest frame.
\emph{Contact to target}: the distance from a contact to its target.
$\Delta$: TouchAtlas $-$ VI smoother (negative is better).
\emph{Objects}: everyday objects. \emph{Coverage}: their far sides touched all over, without targets.
\emph{Targets}: dense pressing on known targets.
Values are in mm. Bold: lower.}
\label{tab:withheld_full}
\end{table*}

\begin{table*}[t]
\centering
\small
\setlength{\tabcolsep}{4pt}\resfit{\textwidth}{\begin{tabular}{@{}lccccc@{}}
\toprule
 & \multicolumn{2}{c}{Hand not detected} & \multicolumn{2}{c}{No visual hand position} & Contact \\
\cmidrule(lr){2-3}\cmidrule(lr){4-5}
Method & in contact & not in contact & in contact & not in contact & to scan \\
\midrule
VI filter & \textbf{85.8}\ci{71.5}{96.3} & \textbf{95.3}\ci{81.3}{114} & \textbf{78.4}\ci{58.7}{90.9} & 81.8\ci{64.7}{94.5} & 24.9\ci{16.6}{36.9} \\
VI smoother & 149\ci{114}{226} & 168\ci{116}{264} & 122\ci{88.4}{158} & 94.0\ci{68.4}{134} & 51.1\ci{25.7}{55.7} \\
HaWoR & 238\ci{195}{293} & 254\ci{219}{299} & 240\ci{196}{287} & 229\ci{195}{267} & -- \\
\rowcolor{oursrow}TouchAtlas (ours) & 86.5\ci{64.0}{120} & 105\ci{80.2}{126} & 79.2\ci{58.0}{104} & \textbf{76.4}\ci{58.7}{96.7} & \textbf{14.4}\ci{6.4}{20.8} \\
\midrule
VI smoother at $t$ & 414\ci{259}{613} & 325\ci{231}{482} & 305\ci{188}{452} & 197\ci{132}{264} & -- \\
\rowcolor{oursrow}TouchAtlas at $t$ & \textbf{105}\ci{81.1}{137} & \textbf{128}\ci{96.6}{169} & \textbf{89.4}\ci{72.0}{121} & \textbf{96.6}\ci{77.0}{127} & -- \\
\midrule
$\Delta$ (paired) & $-$62.4\ci{$-$137}{$-$20.6}\,{\scriptsize($-$42\%)} & -- & $-$43.2\ci{$-$77.4}{$-$12.8}\,{\scriptsize($-$35\%)} & -- & $-$36.7\ci{$-$48.1}{$-$13.4}\,{\scriptsize($-$72\%)} \\
\bottomrule
\end{tabular}
}
\caption{\textbf{Registration under real occlusion}, the hand behind an object.
The hand position error is measured against the glove tag seen by the second camera.
\emph{Hand not detected}: frames in which the head camera detects no hand.
\emph{No visual hand position}: frames in which the head camera gives no hand position.
\emph{Contact to scan}: the distance from a contact to the scan.
\emph{At $t$}: the estimate for the newest frame.
$\Delta$: TouchAtlas $-$ VI smoother (negative is better). Values are in mm.
Bold: lowest in each block.}
\label{tab:real_occlusion}
\end{table*}

\begin{table*}[t]
\centering
\small
\setlength{\tabcolsep}{4pt}\resfit{\textwidth}{\begin{tabular}{@{}lccccc@{}}
\toprule
Time without vision & 0.6\,s & 1\,s & 2\,s & 5\,s & 10\,s \\
\midrule
\multicolumn{6}{@{}l}{\emph{Contact to target}} \\
\quad VI filter & 22.2\ci{16.9}{46.2} & 20.8\ci{15.8}{29.8} & 33.1\ci{20.3}{43.7} & 55.0\ci{38.6}{63.8} & 68.0\ci{56.9}{79.3} \\
\quad VI smoother & 20.3\ci{16.6}{29.6} & 20.0\ci{19.1}{26.9} & 28.7\ci{18.9}{42.8} & 63.2\ci{53.2}{83.0} & 136\ci{98.5}{223} \\
\quad VI + map projection & 19.2\ci{13.0}{23.3} & 19.3\ci{12.3}{27.3} & 22.2\ci{17.6}{42.8} & 63.2\ci{51.2}{83.0} & 136\ci{93.8}{223} \\
\quad TouchAtlas, no slip only & 16.6\ci{13.2}{24.4} & 17.3\ci{11.7}{20.8} & 24.0\ci{17.5}{35.0} & 40.4\ci{26.6}{54.4} & \textbf{49.6}\ci{41.2}{70.8} \\
\quad TouchAtlas, surface only & 17.6\ci{12.8}{20.0} & 18.6\ci{12.9}{21.9} & 30.6\ci{22.3}{42.0} & 85.0\ci{59.9}{120} & 143\ci{93.9}{200} \\
\rowcolor{oursrow}\quad TouchAtlas (ours) & \textbf{14.6}\ci{9.9}{19.3} & \textbf{14.4}\ci{12.0}{20.1} & 22.1\ci{15.5}{30.7} & \textbf{38.5}\ci{27.7}{49.5} & 58.7\ci{35.6}{69.3} \\
\quad $\Delta$ (paired) & $-$5.7\ci{$-$15.0}{$-$1.0}\,{\scriptsize($-$28\%)} & $-$5.6\ci{$-$12.5}{$-$0.3}\,{\scriptsize($-$28\%)} & $-$6.7\ci{$-$19.6}{$+$0.4}\,{\scriptsize($-$23\%)} & $-$24.7\ci{$-$48.1}{$-$13.7}\,{\scriptsize($-$39\%)} & $-$76.8\ci{$-$160}{$-$47.3}\,{\scriptsize($-$57\%)} \\
\midrule
\multicolumn{6}{@{}l}{\emph{Hand in contact, 3D}} \\
\quad VI filter & 12.8\ci{10.8}{15.5} & 13.9\ci{11.1}{18.0} & 21.8\ci{18.3}{34.6} & 61.2\ci{45.0}{67.0} & 68.5\ci{56.4}{77.1} \\
\quad VI smoother & 10.5\ci{8.9}{11.6} & 12.2\ci{10.2}{15.8} & 28.6\ci{20.5}{34.6} & 86.7\ci{53.7}{111} & 165\ci{120}{261} \\
\quad TouchAtlas, no slip only & 11.0\ci{10.0}{12.0} & 12.9\ci{11.7}{14.3} & 19.9\ci{18.1}{23.0} & 38.8\ci{31.6}{52.5} & \textbf{60.1}\ci{44.4}{86.3} \\
\quad TouchAtlas, surface only & 13.2\ci{10.7}{16.9} & 15.6\ci{12.3}{19.2} & 25.2\ci{21.7}{31.5} & 84.8\ci{70.3}{109} & 171\ci{110}{224} \\
\rowcolor{oursrow}\quad TouchAtlas (ours) & 13.6\ci{11.5}{17.5} & 16.4\ci{13.9}{18.4} & 20.6\ci{18.7}{24.5} & \textbf{34.3}\ci{26.3}{47.9} & 62.6\ci{40.8}{84.4} \\
\quad $\Delta$ (paired) & $+$3.2\ci{$+$1.2}{$+$6.8}\,{\scriptsize($+$30\%)} & $+$4.2\ci{$+$1.2}{$+$6.4}\,{\scriptsize($+$35\%)} & $-$8.0\ci{$-$12.6}{$-$0.4}\,{\scriptsize($-$28\%)} & $-$52.4\ci{$-$78.9}{$-$14.6}\,{\scriptsize($-$60\%)} & $-$102\ci{$-$184}{$-$59.8}\,{\scriptsize($-$62\%)} \\
\midrule
\multicolumn{6}{@{}l}{\emph{Hand in contact, along $\mathbf{n}$}} \\
\quad VI filter & 4.1\ci{2.9}{5.6} & 5.5\ci{4.3}{7.5} & 11.8\ci{7.7}{23.2} & 19.3\ci{9.6}{28.8} & 28.0\ci{23.6}{33.8} \\
\quad VI smoother & 3.7\ci{3.0}{4.6} & 3.7\ci{2.7}{5.3} & 12.1\ci{8.9}{19.0} & 33.3\ci{16.1}{59.9} & 94.9\ci{54.6}{137} \\
\quad TouchAtlas, no slip only & 3.4\ci{2.4}{4.6} & 3.8\ci{2.5}{5.3} & \textbf{6.5}\ci{4.4}{8.3} & 14.8\ci{10.1}{23.5} & 20.7\ci{17.3}{39.1} \\
\quad TouchAtlas, surface only & 6.5\ci{4.8}{9.7} & 6.6\ci{4.6}{11.1} & 7.9\ci{5.1}{13.0} & 14.7\ci{9.8}{27.9} & 22.7\ci{15.4}{48.8} \\
\rowcolor{oursrow}\quad TouchAtlas (ours) & 6.3\ci{4.4}{9.6} & 6.9\ci{4.8}{10.5} & 8.7\ci{4.5}{12.6} & \textbf{10.4}\ci{7.0}{13.3} & \textbf{15.1}\ci{11.5}{18.4} \\
\quad $\Delta$ (paired) & $+$2.6\ci{$+$0.8}{$+$5.9}\,{\scriptsize($+$70\%)} & $+$3.1\ci{$+$0.9}{$+$6.7}\,{\scriptsize($+$84\%)} & $-$3.4\ci{$-$12.3}{$+$2.2}\,{\scriptsize($-$28\%)} & $-$23.0\ci{$-$51.8}{$-$4.9}\,{\scriptsize($-$69\%)} & $-$79.9\ci{$-$120}{$-$39.9}\,{\scriptsize($-$84\%)} \\
\midrule
\multicolumn{6}{@{}l}{\emph{Hand in contact, 3D, by the window length $L$ of the estimator}} \\
\quad TouchAtlas, $L = 0.5$\,s & 13.8\ci{11.3}{17.8} & 16.7\ci{13.1}{18.6} & 19.8\ci{17.6}{23.2} & 39.6\ci{27.1}{50.6} & 74.9\ci{50.8}{93.6} \\
\quad TouchAtlas, $L = 1$\,s & 13.4\ci{11.4}{17.4} & 16.5\ci{13.3}{18.4} & 20.2\ci{18.3}{23.8} & 36.3\ci{27.0}{51.7} & 69.1\ci{46.7}{95.2} \\
\quad TouchAtlas, $L = 2$\,s & 13.6\ci{11.5}{17.5} & 16.4\ci{13.9}{18.4} & 20.6\ci{18.7}{24.5} & 34.3\ci{26.3}{47.9} & 62.6\ci{40.8}{84.4} \\
\quad TouchAtlas, $L = 4$\,s & 13.6\ci{11.5}{17.4} & 16.4\ci{13.8}{18.4} & 20.5\ci{18.6}{24.2} & 30.6\ci{22.0}{55.4} & 49.8\ci{40.9}{64.9} \\
\bottomrule
\end{tabular}
}
\caption{\textbf{Ablations over the time without vision.}
Dense pressing on known targets.
\emph{No slip only} and \emph{surface only}: TouchAtlas with one touch constraint alone.
\emph{Contact to target}: the distance from a contact to its target.
\emph{Hand in contact}: the hand position error during contact, in 3D and along the surface normal.
The last block varies the window length $L$ of the estimator. All other rows use $L = \resWindowL$\,s.
$\Delta$: TouchAtlas $-$ VI smoother (negative is better). Values are in mm.
Bold: lowest, and significantly below the VI smoother.}
\label{tab:window}
\end{table*}

\begin{table*}[t]
\begin{minipage}[t]{\columnwidth}\vspace{0pt}
\centering
\small
\footnotesize\setlength{\tabcolsep}{2.5pt}\resfit{\columnwidth}{\begin{tabular}{@{}lccc@{}}
\toprule
Contacts that may create patches & Dist.\ p50 & Angle p50 & Hand \\
\midrule
\rowcolor{oursrow}Hand observed (ours) & \textbf{5.4--13.5} & \textbf{29.9--66.5$^\circ$} & \textbf{24.4}\ci{21.1}{28.8} \\
Also hand not observed & 9.8--39.3 & 46.1--70.7$^\circ$ & 30.4\ci{27.3}{34.9} \\
\midrule
$\Delta$ (paired) & -- & -- & $+$5.9\ci{$+$2.7}{$+$8.8} \\
\bottomrule
\end{tabular}
}
\caption{\textbf{Which contacts may create a touch patch} (\cref{prop:touch}\,(iii)).
TouchAtlas creates patches only from contacts made while the hand is observed. The second row lets all contacts create patches.
\emph{Dist.}\ and \emph{Angle}: the distance and the normal angle of the patches to the scan, on surfaces the head camera did not see.
\emph{Hand}: the hand position error during contact, without a visual hand pose.
$\Delta$: second row minus first row.
Distances and errors are in mm. Bold: lower.}
\label{tab:surface_models}
\vspace{14pt}
\centering
\small
\footnotesize\setlength{\tabcolsep}{2.5pt}\resfit{\columnwidth}{\begin{tabular}{@{}ll@{}}
\toprule
Quantity & Value \\
\midrule
Camera / hand tracking rate & 30.0\,Hz / 25.2--25.8\,Hz \\
Estimator time per update, mean (p95) & 12.6\,ms (40.1\,ms) \\
Latency to final contact & 2.0\,s \\
\bottomrule
\end{tabular}
}
\caption{\textbf{Runtime.}
\emph{Estimator time}: the time of one update.
\emph{Latency}: the time until a contact is final.}
\label{tab:detection}
\end{minipage}\hfill
\begin{minipage}[t]{\columnwidth}\vspace{0pt}
\centering
\footnotesize\setlength{\tabcolsep}{2pt}\resfit{\columnwidth}{\begin{tabular}{@{}clccc@{}}
\toprule
$k$ & Method & 3D & along $\mathbf{n}$ & across $\mathbf{n}$ \\
\midrule
\multirow{3}{*}{1} & VI smoother & 27.2\ci{22.4}{32.4} & 12.9\ci{8.7}{20.9} & \textbf{18.7}\ci{13.1}{25.1} \\
 & \cellcolor{oursrow}TouchAtlas (ours) & \cellcolor{oursrow}\textbf{25.8}\ci{22.4}{30.3} & \cellcolor{oursrow}\textbf{8.5}\ci{4.4}{17.4} & \cellcolor{oursrow}22.0\ci{14.0}{26.5} \\
 & $\Delta$ (paired) & $-$1.4\ci{$-$6.1}{$+$4.4} & $-$4.5\ci{$-$12.0}{$+$1.4} & $+$3.3\ci{$-$5.0}{$+$9.2} \\
\midrule
\multirow{3}{*}{2} & VI smoother & 44.6\ci{30.9}{58.4} & 20.2\ci{16.3}{27.3} & 49.5\ci{27.3}{66.7} \\
 & \cellcolor{oursrow}TouchAtlas (ours) & \cellcolor{oursrow}\textbf{23.7}\ci{18.1}{30.2} & \cellcolor{oursrow}\textbf{9.5}\ci{6.8}{20.2} & \cellcolor{oursrow}\textbf{17.4}\ci{7.8}{24.5} \\
 & $\Delta$ (paired) & $-$20.9\ci{$-$37.8}{$-$7.0} & $-$10.8\ci{$-$16.7}{$+$0.7} & $-$32.1\ci{$-$58.2}{$-$12.3} \\
\midrule
\multirow{3}{*}{3} & VI smoother & 28.6\ci{24.1}{35.9} & \textbf{9.5}\ci{5.6}{14.5} & 18.2\ci{14.0}{29.1} \\
 & \cellcolor{oursrow}TouchAtlas (ours) & \cellcolor{oursrow}\textbf{25.2}\ci{18.4}{32.7} & \cellcolor{oursrow}13.0\ci{7.7}{17.8} & \cellcolor{oursrow}\textbf{16.8}\ci{8.1}{22.8} \\
 & $\Delta$ (paired) & $-$3.4\ci{$-$13.5}{$+$4.2} & $+$3.5\ci{$-$3.3}{$+$9.4} & $-$1.3\ci{$-$14.5}{$+$7.5} \\
\bottomrule
\end{tabular}
}
\caption{\textbf{What touch constrains} (\cref{prop:touch}\,(i)), all components of \cref{tab:k}.
Hand position error during contact with the visual pose withheld, for contacts on $k$ independent normals.
$\Delta$: TouchAtlas $-$ VI smoother (negative is better).
Values are in mm. Bold: lower.}
\label{tab:k_full}
\end{minipage}
\end{table*}

\end{document}